\documentclass[sigconf]{acmart}
\AtBeginDocument{%
  \providecommand\BibTeX{{%
    \normalfont B\kern-0.5em{\scshape i\kern-0.25em b}\kern-0.8em\TeX}}}

\copyrightyear{2024}
\acmYear{2024}
\setcopyright{rightsretained}
\acmConference[WWW '24]{Proceedings of the ACM Web Conference 2024}{May 13--17, 2024}{Singapore, Singapore}
\acmBooktitle{Proceedings of the ACM Web Conference 2024 (WWW '24), May 13--17, 2024, Singapore, Singapore}\acmDOI{10.1145/3589334.3645611}
\acmISBN{979-8-4007-0171-9/24/05}

\makeatletter
\gdef\@copyrightpermission{
  \begin{minipage}{0.3\columnwidth}
   \href{https://creativecommons.org/licenses/by/4.0/}{\includegraphics[width=0.9\textwidth]{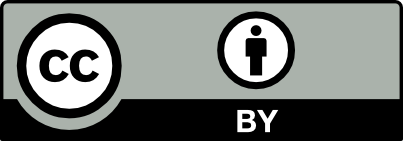}}
  \end{minipage}\hfill
  \begin{minipage}{0.7\columnwidth}
   \href{https://creativecommons.org/licenses/by/4.0/}{This work is licensed under a Creative Commons Attribution International 4.0 License.}
  \end{minipage}
  \vspace{5pt}
}
\makeatother

\usepackage{enumitem}
\usepackage{array,multirow,graphicx}
\usepackage{float}
\usepackage{xcolor,colortbl}

\newcommand{\ie}{\emph{i.e., }}
\newcommand{\eg}{\emph{e.g., }}

\newcommand{\etc}{\emph{etc. }}

\begin{document}

\title{Learning to Generate Explainable Stock Predictions using Self-Reflective Large Language Models}
\vspace{-10px}

\author{Kelvin J.L. Koa}
\affiliation{%
  \institution{National University of Singapore}
  \country{}
  }
\email{kelvin.koa@u.nus.edu}

\author{Yunshan Ma}
\authornote{Corresponding author.}
\affiliation{%
  \institution{National University of Singapore}
  \country{}
  }
\email{yunshan.ma@u.nus.edu}

\author{Ritchie Ng}
\affiliation{%
  \institution{Eastspring Investments, Singapore}
  \country{}
  }
\email{ritchie.ng@eastspring.com}

\author{Tat-Seng Chua}
\affiliation{%
  \institution{National University of Singapore}
  \country{}
  }
\email{dcscts@nus.edu.sg}



\vspace{-10px}
\begin{abstract}
Explaining stock predictions is generally a difficult task for traditional non-generative deep learning models, where explanations are limited to visualizing the attention weights on important texts. Today, Large Language Models (LLMs) present a solution to this problem, given their known capabilities to generate human-readable explanations for their decision-making process. However, the task of stock prediction remains challenging for LLMs, as it requires the ability to weigh the varying impacts of chaotic social texts on stock prices. The problem gets progressively harder with the introduction of the explanation component, which requires LLMs to explain \textit{verbally} why certain factors are more important than the others. On the other hand, to fine-tune LLMs for such a task, one would need expert-annotated samples of explanation for every stock movement in the training set, which is expensive and impractical to scale.  


To tackle these issues, we propose our Summarize-Explain-Predict (SEP) framework, which utilizes a verbal self-reflective agent and Proximal Policy Optimization (PPO) that allow a LLM teach itself how to generate explainable stock predictions, in a fully autonomous manner. The reflective agent learns how to explain past stock movements through a self-reasoning process, while the PPO trainer trains the model to generate the most likely explanations given the input texts at test-time. The training samples for the PPO trainer are also the responses generated during the reflective process, which eliminates the need for human annotators. Using our SEP framework, we fine-tune a specialized LLM that can outperform both traditional deep-learning and LLM methods in prediction accuracy and Matthews correlation coefficient, for the stock classification task. To justify the generalization capability of our framework, we further test it on the portfolio construction task, and demonstrate its effectiveness through various portfolio metrics. Our code can be accessed through https://github.com/koa-fin/sep.

\vspace{-5px}
\end{abstract}

\ccsdesc[500]{Information systems~Web mining}
\ccsdesc[500]{Applied computing~Forecasting}
\ccsdesc[300]{Applied computing~Economics}

\keywords{Stock Prediction, Large Language Models, Explainable AI
\vspace{-2px}
}

\maketitle
\renewcommand{\shortauthors}{Kelvin J.L. Koa, Yunshan Ma, Ritchie Ng, \& Tat-Seng Chua}
\vspace{-5px}
\vspace{-2px}
\section{Introduction} 
The Efficient Market Hypothesis (EMH) states that in financial markets, stock prices reflect all available information \cite{fama1970efficient}, and should only react to new information. Through mining and analysing external data sources, the goal of investors is to quickly understand the impact of new information on the market, in order to anticipate future stock price movements \cite{gidofalvi2001using}. However, analyzing the impact of these data on the stock market is a huge undertaking and imposes a heavy workload on financial experts, due to the large volume of information available \cite{feng2021hybrid}. Because of this, many have explored the use of deep-learning techniques \cite{feng2019temporal, li2021modeling, koa2023diffusion} for stock prediction.

\begin{figure}[ht]
\vspace{-8px}
\includegraphics[width=\columnwidth]{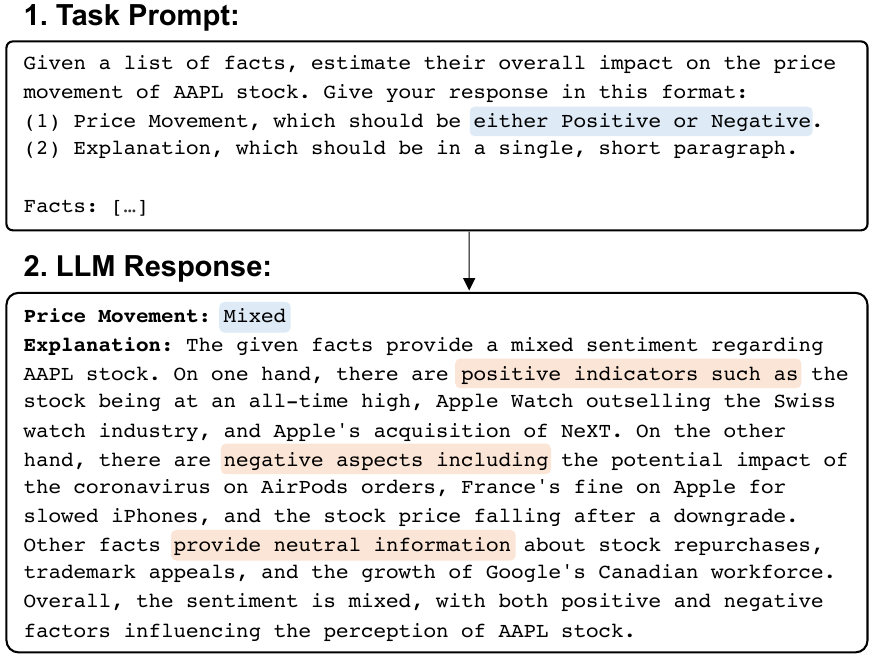}
\vspace{-21px}
\caption{
While LLMs can classify the sentiment of individual texts (highlighted in orange), they are not trained to weigh between them to produce an aggregate prediction (highlighted in blue). An improved response by our fine-tuned LLM will be presented in the results. [...] refers to truncated text.
}
\label{uninformative}
\vspace{-10px}
\end{figure}
However, due to their complex and quantitative nature, traditional deep-learning methods in stock prediction are
black box models and do not address the explainability of their predictions \cite{li2023pen}. This reduces their usability in practical applications, as users might not 
be able to 
trust \cite{biran2017human} the results to invest their capital. Even among 
works that deal with explainable stock predictions \cite{carta2021explainable, li2023pen}, the "explanations" are often 
simply 
defined as the specific texts that caused the price movement, which are usually obtained by analyzing learnt attention weights \cite{deng2019knowledge, sawhney2020deep}. For example, these models could analyze a series of texts regarding Apple stock and determine that its \textit{Positive} prediction is attributed to 
the text 
"\textit{Apple reported 
revenue of \$90.1 billion, beating expectations}". 
However, these models do not go beyond that to explain \textit{why} these texts caused the stock movement, and require the user to make their own inference.

Today, the emergence of Large Language Models (LLMs) has presented a solution to this problem. Recent surveys \cite{yang2023harnessing, zhao2023survey} have shown that LLMs possess both strong Natural-Language Understanding capabilities, which allow them to perform tasks like text summarization \cite{pu2023summarization} and text classification \cite{liang2022holistic} in a few-shot manner; and strong Natural-Language Generation capabilities, which let them 
generate human-readable explanations for their own decision-making process \cite{liu2023chain, wei2022chain}. Currently, works that utilize LLMs for stock prediction \cite{yu2023temporal, chen2023chatgpt} are few, and use limited techniques such as pre-trained LLMs or instruction tuning. Our work seeks to fill this gap by designing a reinforcement learning (RL) framework which can fine-tune a LLM to generate 
explanations for 
stock prediction.

To tackle the explainable stock prediction task using LLMs, we can identify two main challenges. Firstly, it is well-established in past stock prediction literature that social texts are \textit{chaotic}, where the influence of different texts on stock prices can be highly diverse \cite{hu2018listening, xu2018stock}. For example, breaking news such as surprise earnings 
announcements or crisis events often have a visible impact on the stock price, while unsubstantiated opinions or vague remarks usually cause little to no change \cite{sprenger2011news}. This requires a prediction model to have the ability to weigh the varying impacts of new information \cite{fama2015five}, and arrive at a maximum-likelihood prediction \cite{giglio2022factor}. Typically, this involves training a regression-based neural network, and is not a known capability of LLMs (see Figure \ref{uninformative}). Secondly, the problem becomes progressively harder with the introduction of the explanation component, as it requires the LLM to explain \textit{verbally} why certain information are more important than others. However, to train a LLM for this task using RL \cite{ouyang2022training, hu2021lora}, one would need 
good and bad samples \cite{lee2023rlaif, liu2023chain} of explanations for each price movement in the training set. This requires substantial amount of labour by financial experts, which is expensive and impractical to scale. 

\begin{figure*}[h]
\vspace{-5px}
\includegraphics[width=\textwidth]{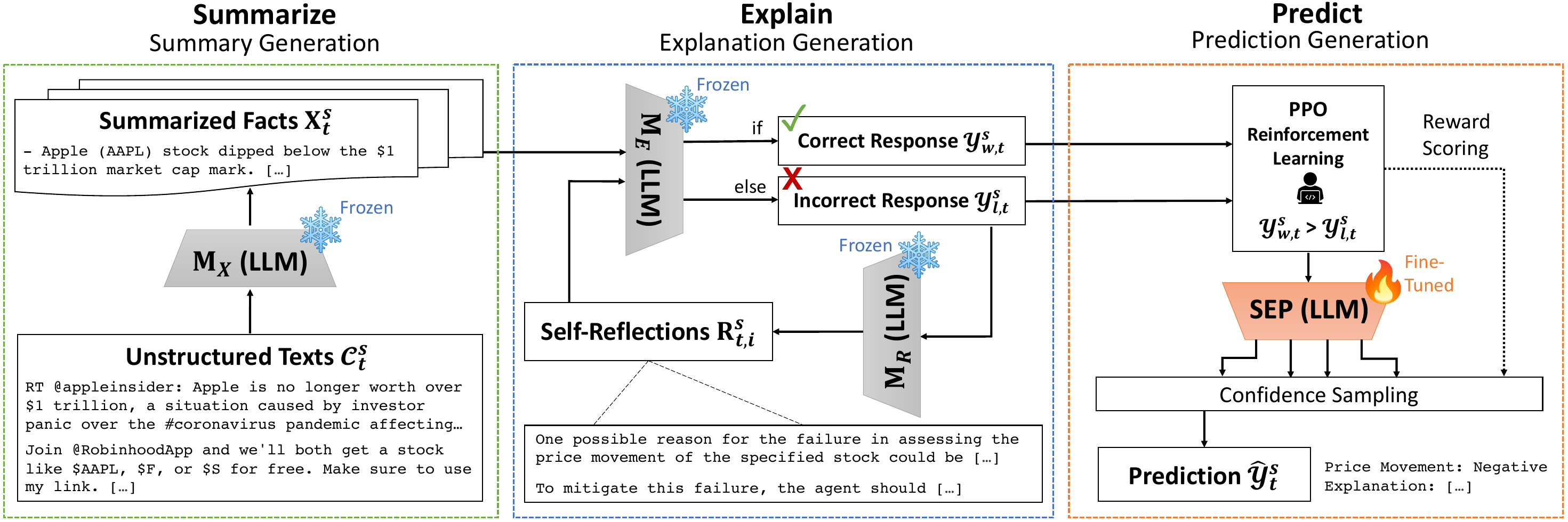}
\vspace{-18px}
\caption{Overall framework of our proposed SEP method, which consists of three components: Summarize, Explain and Predict.}
\label{sep-model}
\vspace{-12px}
\end{figure*}

To deal with the above-mentioned problems, we propose our Summarize-Explain-Predict (SEP) framework, which utilizes a self-reflective agent \cite{shinn2023reflexion} and Proximal Policy Optimization (PPO) \cite{schulman2017proximal} to let a LLM teach itself how to make explainable stock predictions in a fully autonomous manner (see Figure \ref{sep-model}). Firstly, the Summarize module utilizes the strong summarization capabilities of LLMs \cite{pu2023summarization} to convert large volumes of text input data into point-form summaries of factual information. Secondly, in the Explain module, a reflective agent teaches itself to generate correct stock predictions and explain their reasoning \cite{wei2022chain} given a sequence of summarized facts, via an iterative, verbal self-reflective process \cite{madaan2023self, shinn2023reflexion}. The iterative process additionally allows us to obtain a series of \textit{correct} and \textit{incorrect} predictions with annotated explanations through its past mistakes, which can be used as fine-tuning samples without human-in-the-loop. Lastly, in the Predict module, a specialized LLM is fine-tuned \cite{ouyang2022training, hu2021lora} via PPO training \cite{schulman2017proximal} using its own self-taught responses, in order to generate the most likely stock predictions and explanations, given the input texts from an unseen test set.
  
To demonstrate the effectiveness of the SEP framework, we validate through experimental results that our model is able to outperform both traditional deep-learning and LLM methods in terms of its prediction accuracy and Matthews correlation coefficient (MCC) for the binary stock classification task. We also analyze some responses from the fine-tuned LLM qualitatively, to show how it is better able to understand and weigh the impacts of different information within the input texts. Additionally, to justify the generalization capability of the framework, we test
it on the portfolio construction task, by generating explainable weights for a stock portfolio. We also demonstrate the effectiveness of this method through portfolio metrics, such as its profitability and Sharpe Ratio. 

The main contributions of this paper are summarized as:
\begin{itemize}[leftmargin=*]
\vspace{-2px}
\item We investigate the limitations of teaching LLMs to weigh information in multiple texts for stock prediction in an explainable manner, without expert-annotated explanation samples.

\item We propose a solution that utilizes a self-reflective agent and PPO techniques, that can allow a LLM to teach itself how to make explainable stock predictions in a fully autonomous manner.

\item We validate the effectiveness of SEP through experimental results on tweet data, and show that the fine-tuned LLM is able to provide improvements in both the prediction performance and the quality of its explanations. We further demonstrate the generalizability of the framework by fine-tuning a LLM to generate quantitative weights for multiple stocks, to tackle the portfolio task.
\end{itemize}
\vspace{-10px}
\section{Related Works}
In this section, we trace the progress of textual analysis techniques in stock prediction works, 
and also explore some pioneering works that utilized Large Language Models (LLMs) in the financial domain.

\textbf{Text Analysis in Stock Prediction.}
Early text analysis works in stock prediction first studied the effectiveness of using different textual representations of news, such as Bag of Words, Noun Phrases, and Named Entities, in Support Vector Machines (SVM) \cite{schumaker2009textual}. These "shallow" features were later replaced in favor of structured information, where events in the form of (\textit{Actor, Action, Object}) tuples were used as inputs for deep neural networks \cite{ding2014using, ding2015deep}. 

Later works would define the challenges in text analysis more clearly, which was attributed to the chaotic and diverse nature of text data \cite{hu2018listening}. This led to the popular use of attention-based models to capture the "most important" information in texts directly from pre-trained text embeddings \cite{deng2019knowledge}. Some other notable works include the use of Variational Auto-Encoders (VAEs) to model the latent factors in market information \cite{xu2018stock}, and Transformer models \cite{yang2020html}. 

Most recent works have moved away from improving text analysis methods, and opted instead to enhance the current models with additional forms of information, such as the vocal features from audio data \cite{yang2022numhtml} or cross-stock 
 impacts from company relational graphs \cite{sawhney2020deep, li2021modeling}. In contrast, our work return to purely text-based models, to isolate the effects of text information on stock movements.

\textbf{Large Language Models in Finance.}
Out of the existing works that utilize LLMs on general financial tasks, the most well-known one is BloombergGPT \cite{wu2023bloomberggpt}, 
which trained a 50B parameters LLM using their existing large financial text corpus. Their model was evaluated on several downstream tasks such as sentiment analysis and named-entity recognition (NER), with optimistic results. Along this direction, some works have also attempted to fine-tune their own financial LLM, which include FinMA \cite{xie2023pixiu} and FinGPT \cite{yang2023fingpt}. 

Other works explored the use of existing LLMs such as ChatGPT to perform specialized downstream tasks, such as stock sentiment prediction from news headlines \cite{lopez2023can}, and classification of Federal announcements \cite{hansen2023can}. These early works focused on analyzing \textit{individual} texts, as opposed to a sequence of texts. More recent works have explored the use of LLMs to make stock predictions using sequences of stock-related texts, via instruction-tuning \cite{yu2023temporal} or pre-trained models enhanced with relational graphs \cite{chen2023chatgpt}. We build on these works by implementing an additional verbal self-reflective agent to learn how to generate better explanations, and a PPO trainer to fine-tune a more specialized LLM for stock predictions.
\vspace{-5px}
\section{Methodology}
In this section, we first define the task and data for explainable stock prediction. We then present the proposed SEP framework, which was illustrated in Figure \ref{sep-model}. There are three main components: (1) a Summarize module, which generates a summary of factual information from the unstructured text inputs; (2) an Explain module, which generates explanations for its stock predictions and refines them through an iterative self-reflective process; and (3) a Predict module, which generates confidence-based predictions after fine-tuning a LLM using its self-generated annotated samples.

\vspace{-5px}
\subsection{Preliminaries}
\subsubsection{Problem Formulation}
Given a stock $s \in \mathcal{S}=\left\{s_i\right\}_{i=1}^{O}$ and its associated text corpora for the past $T$ days $\left\{\mathbf{\mathcal{C}}^{s}_{t-T}, \cdots, \mathbf{\mathcal{C}}^{s}_{t-2}, \mathbf{\mathcal{C}}^{s}_{t-1}\right\}$, we aim to generate a stock prediction for the next trading day $\hat{\mathbf{\mathcal{Y}}}^{s}_{t}$, which consists of a binary price movement $\hat{\mathbf{y}}^{s}_{t} \in \{0, 1\}$ and a human-readable explanation $\hat{\mathbf{e}}^{s}_{t}$. 
Each corpus contains a variable number of unstructured texts
$\mathcal{C}_{t}^{s} = \left\{\mathbf{c}^{s}_{t, n} \right\}_{n=1}^{N_t^s}$,
where $\mathbf{c}^{s}_{t, n}$ is a single text, and $N_t^s =\lvert \mathcal{C}_{t}^{s} \rvert$ is the number of texts for the stock $s$ on day $t$. 

\vspace{-2px}
\subsubsection{Data Collection and Clustering}
In this work, we construct a new dataset by following the data collection methodology used for the \textbf{ACL18} StockNet dataset \cite{xu2018stock}, which is a popular benchmark used in many stock prediction works \cite{feng2018enhancing, sawhney2020deep, li2023pen}. The duration of the original dataset ranges from year 2014–2016, and we collect an updated version for year 2020–2022. Since the previous work, the number of industries have expanded, and the number of tweets have also increased exponentially. We collect data for the top 5 stocks in the 11 industries, giving us a total of 55 stocks. The price data is collected from Yahoo Finance\footnotemark{}\footnotetext{https://finance.yahoo.com/}, while the tweet data is collected using the Twitter API\footnotemark{}\footnotetext{https://developer.twitter.com/}. Additionally, given the large volume of tweets for each day, 
we utilize a clustering pipeline via BERTopic \cite{grootendorst2022bertopic} to identify the representative tweets for each day. These tweets would be used as the text inputs for all models. More details on the dataset and clustering pipeline can be found in Appendix \ref{datasets}. 

\vspace{-5px}
\subsection{Summary Generation}
The goal of the Summary module is to generate summarized information from the unstructured input texts. 
Current LLMs are known for their summarization ability, which surpass even humans \cite{pu2023summarization}.
Given that a sequence of raw texts from $T$ days would exceed the character limit, even for 16K-context LLMs, 
we first prompt a LLM to generate point-form summaries of factual information from the given texts \cite{yu2023temporal} for each day. The prompt takes in two variable inputs: the specified stock $s$, and the unstructured text inputs $\mathbf{\mathcal{C}}_{t}^{s}$ for each day $t$. The LLM $\text{M}_{X}$ then generates a summary of facts $\mathbf{X}_{t}^{s}$ that can be learnt from the input texts, which can include specific information for stock $s$ and other related news in its industry, \eg "\textit{Big Tech stocks, including Apple (AAPL), Google, Amazon, and Facebook, beat earnings expectations.}" This can be formulated as:
\begin{equation}
\mathbf{X}_{t}^{s} = \text{M}_{X}\left(s, \mathbf{\mathcal{C}}_{t}^{s}\right).
\label{extractor}
\end{equation}

Within the prompt, we also provide two in-context examples \cite{ye2023context} that were composed from selected cases in the dataset. 
Full examples for all prompts in this work can be found in Appendix \ref{prompt_samples}.

\vspace{-5px}
\subsection{Explanation Generation}
The goal of the Explain module is two-fold: While the key aim of the module is to generate clear explanations for stock predictions, the generated explanations also serve as a reasoning step \cite{wei2022chain} for the LLM to do self-reflection to improve its own predictions \cite{shinn2023reflexion}. In the following subsections, we discuss the initial prompt design and the subsequent self-reflective process for the module.

\vspace{-2px}
\subsubsection{Explanation Prompting}
The prompt for the Explain module contains two variable inputs: the specified stock $s$, and a \textit{sequence} of extracted information that was generated from the previous module. 
Given these inputs, the LLM $\text{M}_{E}$ then generate the response $\mathbf{\mathcal{Y}}^{s}_{t}$, which should contain the next-day price movement $\mathbf{y}_{t}^{s}$, and the annotated explanation $\mathbf{e}_{t}^{s}$, \ie $\mathbf{\mathcal{Y}}^{s}_{t} = (\mathbf{y}_{t}^{s}, \mathbf{e}_{t}^{s})$. We formalize this as:
\begin{equation}
\vspace{-2px}
\mathbf{\mathcal{Y}}^{s}_{t} = \text{M}_{E}\left(s, \mathbf{X}_{t-T}^{s}, \cdots, \mathbf{X}_{t-2}^{s}, \mathbf{X}_{t-1}^{s}\right).
\label{explainer}
\end{equation}

Similar to the previous summarization prompt, we select two cases from the dataset and manually compose the response trajectories to use as few-shot exemplars \cite{ye2023context}. Additionally, the two example cases chosen have specifically one Positive and one Negative movement label, in order to avoid any majority label bias \cite{zhao2021calibrate}. The prompt trajectories are designed in a fashion similar to ReAct \cite{yao2022react}, albeit in a singular, prediction-explanation step. 

\vspace{-2px}
\subsubsection{Self-Reflective Process}
\begin{figure}[h]
\vspace{-10px}
\includegraphics[width=\columnwidth]{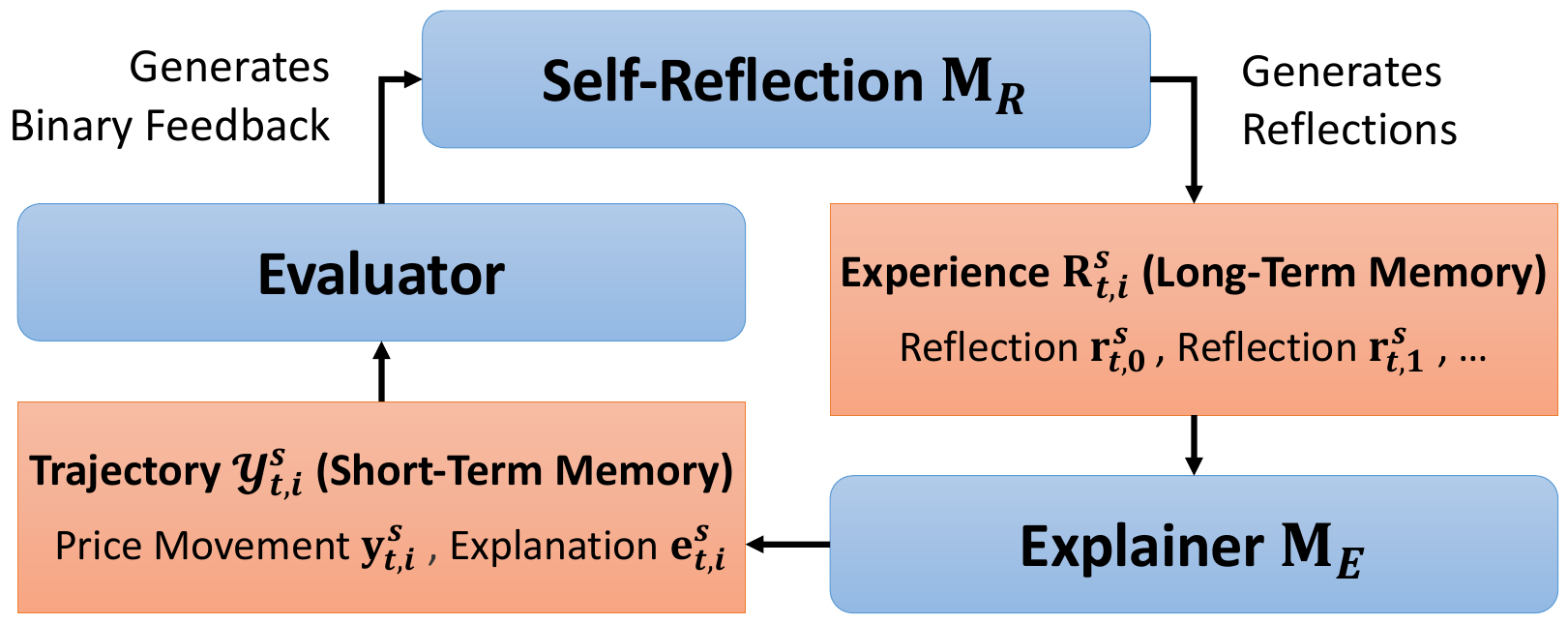}
\vspace{-20px}
\caption{Diagram of the self-reflective process.}
\label{reflection}
\vspace{-10px}
\end{figure}

Current LLMs are not trained to generate stock predictions, which could cause incorrectly-generated annotated examples in the previous step. To tackle this, we deploy the LLM as an autonomous agent that can iteratively improve on its past responses, through a verbal self-reflection loop (see Figure \ref{reflection}). 
The loop is first seeded with the response from the previous step, \ie $\mathbf{\mathcal{Y}}^{s}_{t, 0} = \mathbf{\mathcal{Y}}^{s}_{t}$, which is taken to be the initial iteration $i=0$. 

From the generated price movement $\mathbf{y}_{t, i}^{s}$, we can obtain a binary feedback by evaluating its alignment with the ground truth. For the incorrect samples, we then prompt a LLM $\text{M}_{R}$ to generate a verbal feedback $\mathbf{r}_{t, i}^{s}$ for each iteration $i$, given its previous inputs and outputs, which we refer to as its short-term memory \cite{shinn2023reflexion}. The feedback should explain clearly where it went wrong in its previous reasoning $\mathbf{e}_{t, i}^{s}$, and also come up with a high-level plan to mitigate this failure for the next iteration. The overall formalization is:
\vspace{-1px}
\begin{equation}
\vspace{-1px}
\mathbf{r}_{t, i}^{s} = \text{M}_{R}\left(s, \mathbf{X}_{t-T}^{s}, \cdots, \mathbf{X}_{t-2}^{s}, \mathbf{X}_{t-1}^{s}, \mathbf{\mathcal{Y}}^{s}_{t, i}\right).
\label{reflection-eqn}
\vspace{-2px}
\end{equation}

For every iteration, each reflection $\mathbf{r}_{t, i}^{s}$ represent a lesson that the LLM learnt from its failures, which is added to its experiences, or long-term memory \cite{shinn2023reflexion}. We represent this as a set of reflections, 
$\mathbf{R}_{t, i}^{s} = \left[\mathbf{r}_{t, 0}^{s}, \mathbf{r}_{t, 1}^{s}, \cdots, \mathbf{r}_{t, i}^{s}\right]$. 
The reflections, together with the original inputs, are fed again into LLM $\text{M}_{E}$ to generate the price movement and explanation for the next iteration. The formalization is:
\vspace{-2px}
\begin{equation}
\vspace{-1px}
\mathbf{\mathcal{Y}}^{s}_{t, i} = \text{M}_{E}\left(s, \mathbf{X}_{t-T}^{s}, \cdots, \mathbf{X}_{t-2}^{s}, \mathbf{X}_{t-1}^{s}, \mathbf{R}_{t, i}^{s}\right).
\label{explainer2}
\vspace{-2px}
\end{equation}
The prompt and response examples can be found in Appendix \ref{prompt_samples}. 

Through this process, we are then able to obtain pairs of correct and incorrect responses, for each successful reflection. We define these as $\mathbf{\mathcal{Y}}_{w, t}^{s} = \left(\mathbf{y}_{t, \tilde{i}}^{s}, \mathbf{e}_{t, \tilde{i}}^{s}\right)$ and $\mathbf{\mathcal{Y}}_{l, t}^{s} = \left(\mathbf{y}_{t, \tilde{i}-1}^{s}, \mathbf{e}_{t, \tilde{i}-1}^{s}\right)$ respectively, where $\tilde{i}$ refers to the iteration in which the reflective process resulted in the LLM $\text{M}_{E}$ generating the correct stock movement. 

\subsection{Prediction Generation}
\vspace{-1px}
The goal of the Predict module is to fine-tune a LLM to generate good stock predictions and explanations for the unseen test period. In this section, we discuss the overall fine-tuning process of the model and the subsequent inference procedure at test-time. 

\vspace{-5px}
\subsubsection{Model Fine-Tuning} 
\begin{figure}[h]
\vspace{-12px}
\includegraphics[width=\columnwidth]{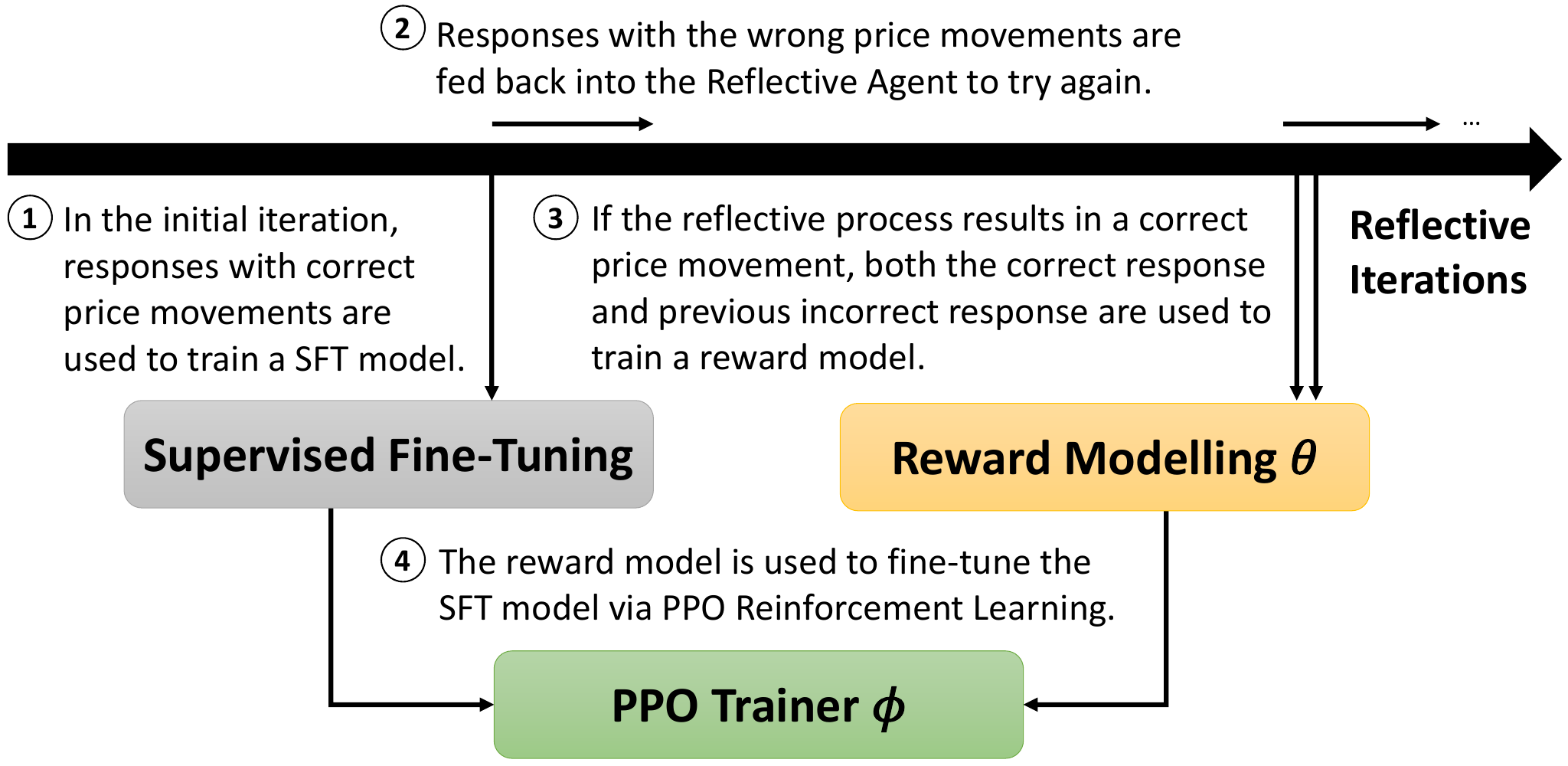}
\vspace{-20px}
\caption{Diagram of the fine-tuning process.}
\label{ppo}
\vspace{-12px}
\end{figure}

Following previous works that tackles Reinforcement Learning from Human Feedback (RLHF) \cite{ouyang2022training, stiennon2020learning}, we utilize a similar three-step process to fine-tune a LLM. Instead of human feedback, we use the binary evaluations from the reflections to choose the "better" response during training (see Figure \ref{ppo}). 

In the first step, we collect the demonstration data, which are taken from the correct predictions in the initial iteration $\mathbf{\mathcal{Y}}^{s}_{t, 0}$. These samples do not have corresponding "wrong" responses, as they were taken from the initial prompt. The samples are used to train a supervised policy $\pi^{SFT}$ using Supervised Fine-Tuning (SFT).

In the second step, we collect the comparison data $\mathcal{D}$, which contains pairwise correct and incorrect responses $\mathbf{\mathcal{Y}}_{w, t}^{s}, \mathbf{\mathcal{Y}}_{l, t}^{s}$ for each structured input $\mathbf{X}^{s}_{t}$, taken from the successful reflection iterations. These are used to train a reward model $r_{\theta}$, which learns to give higher reward scores to the correct responses. Specifically, we train the model to minimize the following cross-entropy loss \cite{stiennon2020learning}:
\vspace{-2px}
\begin{equation}
\vspace{-1px}
\mathcal{L}(\boldsymbol{\theta})=-\mathbb{E}_{\left(\mathbf{X}, \mathbf{\mathcal{Y}}_{w}, \mathbf{\mathcal{Y}}_{l}, s, t\right)\sim \mathcal{D}}\left[\textup{log}\left(\sigma\left(r_{\theta}\left(\mathbf{X}_{t}^{s}, \mathbf{\mathcal{Y}}_{w, t}^{s}\right)-r_{\theta}\left(\mathbf{X}_{t}^{s},\mathbf{\mathcal{Y}}_{l, t}^{s}\right)\right)\right)\right].
\label{reward_loss}
\end{equation}

In the third step, we use the reward model to optimize the trained policy using PPO \cite{schulman2017proximal}. We first initialize the model with the supervised policy $\pi^{SFT}$, and use it to generate predictions $\mathbf{\hat{\mathcal{Y}}}^{s}_{t}$ for randomly selected samples $\mathbf{X}^{s}_{t}$ from the overall dataset $\mathcal{D}_{\pi_{\phi}^{RL}}$. Next, the reward model $r_{\theta}$ is used to generate a reward for each response. We then try to optimize a PPO model $\pi_{\phi}^{RL}$ by maximizing the overall reward. This is achieved by minimizing the following loss objective:
\vspace{-8px}
\begin{equation}
\vspace{-2px}
\mathcal{L}(\boldsymbol{\phi})=-\mathbb{E}_{\left(\mathbf{X}, \mathbf{\hat{\mathcal{Y}}}, s, t\right)\sim \mathcal{D}_{\pi_{\phi}^{RL}}}\left [  r_{\theta}\left(\mathbf{X}_{t}^{s}, \mathbf{\hat{\mathcal{Y}}}_{t}^{s}\right)-\beta \textup{log} \frac{\pi_{\phi}^{RL}\left(\mathbf{\hat{\mathcal{Y}}^{s}}_{t}|\mathbf{X}^{s}_{t}\right)}{\pi^{SFT}\left(\mathbf{\hat{\mathcal{Y}}^{s}}_{t}|\mathbf{X}^{s}_{t}\right)}\right ].    
\label{ppo_loss}
\end{equation}

We note that the objective includes an additional term that penalizes the KL divergence between the trained policy $\pi_{\phi}^{RL}$ and the supervised policy $\pi^{SFT}$ \cite{jaques2019way}, which is used to deter the policy from collapsing into a single mode \cite{stiennon2020learning}, and prevent it from generating responses that are too different from those of the original reference model $\pi^{SFT}$ \cite{yao2023retroformer}. The term is controlled by a hyper-parameter $\beta$.

\vspace{-2px}
\subsubsection{Confidence-based Sampling} During inference, the unstructured input texts $\mathbf{\mathcal{C}}_{t}^{s}$ are first summarized using a pre-trained LLM. We then use the trained policy $\pi_{\phi}^{RL}$ to generate the next-day predictions $\mathbf{\hat{\mathcal{Y}}^{s}}_{t}$ from the summarized facts $\mathbf{X}_{t}^{s}$. For generating predictions, we use a best-of-$n$ sampler, where we generate $n$ responses and use the scores from reward model $r_{\theta}$ to select the best response \cite{yao2023retroformer}.

\vspace{-5px}
\section{Experiment}
\vspace{-2px}
We evaluate the performance of SEP on our collected dataset.
Our work aims to answer the following three research questions:
\begin{itemize}[leftmargin=*]
\vspace{-2px}
\item \textbf{RQ1:} How does the SEP model perform against traditional deep-learning and other LLM methods in the stock prediction task, in both its classification accuracy and quality of explanations? 

\item \textbf{RQ2:} How does each proposed component (\ie Summarize, Explain, Predict) help to improve the performance of the SEP model? 

\item \textbf{RQ3:} Is the SEP framework sufficiently generalizable to other finance-related tasks, such as explainable portfolio construction?
\end{itemize}

\vspace{-10px}
\subsection{Experimental Settings}
\vspace{-2px}
\subsubsection{Baselines}
To demonstrate the effectiveness of our SEP-trained model, we compare it against baselines from both traditional deep-learning models and fine-tuned Large Language Models (LLMs). 

\noindent Deep Learning Models:
\begin{itemize}[leftmargin=*]
\vspace{-2px}
\item \textbf{VAE$+$Attention} \cite{xu2018stock}: In this model, a Variational Auto-encoder (VAE) \cite{kingma2013auto} is used to model the latent market factors within texts. News-level \cite{hu2018listening} and temporal \cite{ding2021leveraging} attention are used to weigh texts on their salience in the corpus and across the input period. Texts are represented on the word level using GloVe \cite{pennington2014glove}.

\item \textbf{GRU$+$Attention} \cite{sawhney2020deep}: This model utilize a hierarchical attention model using Gated Recurrent Networks (GRU) \cite{qin2017dual} with multiple stages of attention layers \cite{yang2016hierarchical, bahdanau2014neural} to capture the corpus-level and day-level importance of each text. The texts are encoded on the sentence level using the Universal Sentence Encoder \cite{cer2018universal}. 

\item \textbf{Transformer} \cite{yang2022numhtml}: This model uses stacked transformer encoders to perform multi-headed self-attention on the token- and sentence-level, 
before decoding with multiple feed-forward layers \cite{yang2020html}.
For preprocessing, the texts are encoded on the token level using the Whole Word Masking BERT (WWM-BERT) \cite{devlin2018bert}. 
\end{itemize}

\vspace{-2px}
\noindent Large Language Models:
\begin{itemize}[leftmargin=*]
\vspace{-2px}
\item \textbf{GPT-3.5-turbo} \cite{ouyang2022training}: 
We provide the same prompts to a GPT-3.5-turbo-16k LLM for comparison. ChatGPT has previously been explored in other stock sentiment prediction works \cite{lopez2023can, yu2023temporal}.

\item \textbf{Vicuna-7b-v1.5} \cite{chiang2023vicuna}: Similarly, we provide the same prompts to a Vicuna-7b-v1.5-16k LLM. This is also the model used for fine-tuning in our work, and serves as a base model for comparison.

\item \textbf{FinGPT-Forecaster} \cite{yang2023fingpt}: This is an instruction-tuned LLM model by FinGPT, which can take in a series of market news to make stock predictions. This is the most recent model for our task. 
\end{itemize}

\vspace{-2px}
For the deep-learning methods, we keep only the text-processing components for an equivalent comparison. 
The inputs for all models are the unstructured representative tweets $\mathbf{\mathcal{C}}_{t}^{s}$. 
Following the previous works that deals with the binary stock classification task \cite{xu2018stock, ding2015deep, feng2018enhancing}, we use the prediction accuracy and 
Matthews Correlation Coefficient (MCC) 
as our evaluation metrics. For all LLM results, any predictions that are made in the wrong format, or are "Neutral" or "Mixed", will be considered as an incorrect prediction. 

Additionally, a key feature of the SEP framework is the Summarize module, which extracts key information from unstructured tweets for the LLM to base its predictions on. However, there are some days when there are no useful information to be found in the tweets. In such cases, there can still be significant price movements, which could be due to external factors such as stock price stochasticity \cite{koa2023diffusion} or daily interest rates fluctuations \cite{alam2009relationship}. For the LLM experiments, we report both the results before and after removing such cases. In practice, this could be seen as a benefit of LLMs, as it is able to actively tell that it has not enough information to make a prediction, and investors could choose to either look for more information to analyze or not invest their capital for the day.

\vspace{-5px}
\subsubsection{Implementation Details}
For the Summarize and Explain components, we evaluate two different models for generating the responses. We use OpenAI GPT-3.5-turbo-16k for the top 1 stock in each industry, and Vicuna-13b-v1.5-16k for the remaining stocks. 
Both are set to a temperature of zero.
The input length is $T=5$.

For training the prediction model, we use Vicuna-7b-v1.5-16k.
The LLM is trained using \textit{trl}, which supports transformer reinforcement learning with PPO trainer\footnotemark{}\footnotetext{https://huggingface.co/docs/trl}. For the supervised fine-tuning, we run two epochs with a learning rate of $3\times 10^{-4}$ For the reward model tuning, we run one epoch with a learning rate of $2\times 10^{-4}$. For the RL learning with PPO, we run four epochs with a learning rate of $1.4\times 10^{-5}$. All components are trained using 4-bit quantized low-rank adapters (LoRA) \cite{hu2021lora} with a setting of $r=8$. At inference, we set $n=4$ for $n$-shot sampling, where the temperature of the model is set at 0.7. The best response, based on reward scoring, will be used as the selected output for the final comparisons. 


\vspace{-7px}
\subsection{Performance Comparison (RQ1)}
In this section, we evaluate both the prediction and explanation responses generated by our SEP model, through quantitative and qualitative comparisons against the relevant baselines.

\vspace{-5px}
\subsubsection{Prediction Performance}
\begin{figure}[h!]
\includegraphics[width=\columnwidth]{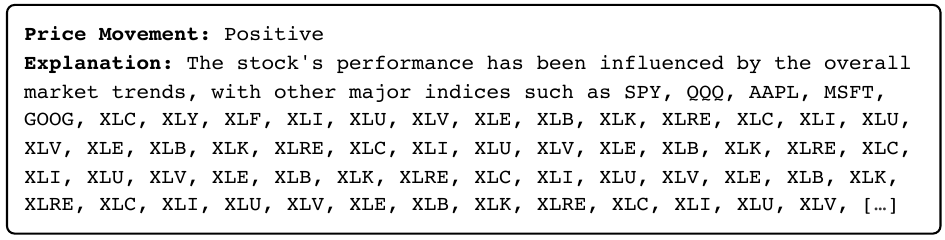}
\vspace{-23px}
\caption{An example of a hallucinated response from Vicuna. [...] refers to truncated text, which are all repeated text.}
\label{hallucinationx}
\end{figure}

Table \ref{results} reports the quantitative results on the stock prediction task. On the prediction accuracy, we observe that the SEP model fine-tuned on the GPT-generated explanations (Table \ref{results}, left) was able to obtain the best results, achieving an improvement of 2.4\% over the strongest baseline (GRU$+$Att) using all texts. On the other hand, the SEP model fine-tuned on explanations generated by Vicuna-v1.5 (Table \ref{results}, right) under-performed the baselines in terms of accuracy. A possible reason for this is that the Vicuna-generated explanations used for training the model are prone to hallucinations, which could negatively impact the reasoning ability of the SEP model (see Figure \ref{hallucinationx}). 
The poorer performance of GPT-3.5, a pre-trained LLM, is largely attributed to its inability to make decisive predictions from mixed sentiments. The instruction-tuned FinGPT-Forecaster is able to improve on this by guiding the LLM towards trained responses, which are in the correct format. Finally, our SEP model produces the best accuracy, likely due to its additional self-reflective process and reinforcement learning.

\begin{table*}[h]
\vspace{-5px}
\caption{Performance comparisons in accuracy and MCC of our SEP model against baselines. The best results are boldfaced.}
\label{results}
\vspace{-10px}
\begin{tabular}{llcccccccc}
\toprule
\multicolumn{2}{c}{\multirow{3}{*}{\textbf{Models}}}                                                             & \multicolumn{4}{c}{Top 1 Stock, GPT-3.5}                                   & \multicolumn{4}{c}{Remaining Stocks, Vicuna}                                          \\ \cmidrule(lr){3-6}  \cmidrule(lr){7-10}
\multicolumn{2}{c}{}                                                                                    & \multicolumn{2}{c}{All Texts}      & \multicolumn{2}{c}{Informative Texts} & \multicolumn{2}{c}{All Texts}                 & \multicolumn{2}{c}{Informative Texts} \\ \cmidrule(lr){3-4}  \cmidrule(lr){5-6} \cmidrule(lr){7-8} \cmidrule(lr){9-10}
\multicolumn{2}{c}{}                                                                                    & \textbf{Accuracy} & \textbf{MCC}   & \textbf{Accuracy}   & \textbf{MCC}    & \textbf{Accuracy} & \textbf{MCC}              & \textbf{Accuracy}   & \textbf{MCC}    \\ \midrule
\multirow{3}{*}{\begin{tabular}[c]{@{}l@{}}Deep-Learning\\ Models\end{tabular}}   & VAE+Att             & 49.96             & 0.0046          & -                   & -               & 49.83             & \multicolumn{1}{l}{0.0070} & -                   & -               \\
                                                                                  & GRU+Att             & 50.15             & 0.0125          & -                   & -               & \textbf{50.77}    & \multicolumn{1}{l}{0.0189} & -                   & -               \\
                                                                                  & Transformer         & 50.06             & 0.0089          & -                   & -               & 50.17             & \multicolumn{1}{l}{0.0135} & -                   & -               \\ \midrule
\multirow{3}{*}{\begin{tabular}[c]{@{}l@{}}Large Language \\ Models\end{tabular}} & GPT-3.5             & 20.80             & 0.0094          & 29.35               & 0.0298           & 17.57             & 0.0027                     & 22.99               & 0.0052           \\
                                                                                  & Vicuna              & 40.85             & 0.0114          & 45.29               & 0.0368           & 39.66             & 0.0115                     & 43.30               & 0.0301           \\
                                                                                  & FinGPT              & 47.61             & 0.0158          & 51.56               & 0.0384           & 45.76             & 0.0161                    & 46.12               & 0.0379	           \\
                                                                                  & SEP (Ours) & \textbf{51.38}    & \textbf{0.0302} & \textbf{54.35}      & \textbf{0.0993}  & 47.59             & \textbf{0.0203}            & \textbf{50.57}               & \textbf{0.0508}  \\ \bottomrule
\end{tabular}
\vspace{-12px}
\end{table*}

For this task, a more telling metric is the Matthews Correlation Coefficient (MCC), which takes into account the ratios of True and False Positives and Negatives of the predictions \cite{chicco2020advantages, chicco2021matthews}. Given that not all stock movements are necessarily caused by the provided texts, the accuracy results might not be fully indicative of the model's natural language processing capabilities, as it includes some random guesses on the non-informative texts. After filtering for informative texts only, we can see increases in the MCC ratio, 
possibly from having 
less random guesses in the prediction results. 

On the MCC metric, our SEP model was able to outperform all models under all settings, which showcase the true ability of the model to understand the impacts of natural language texts on stock movements, after accounting for the random guesses. Under the all-texts setting, we are able to outperform the strongest deep-learning baseline (GRU$+$Att) by 0.0177 for the GPT-3.5-based model, and 0.0014 for the Vicuna-based model. 
After filtering for informative texts only, our fine-tuned SEP model is also able to outperform the strongest LLM baseline, FinGPT-Forecaster, by 0.0609 and 0.0129 for the GPT-3.5 and Vicuna-based SEP models respectively.


\vspace{-8px}
\vspace{-14px}

\subsubsection{Explanation Performance}
In addition to generating better predictions, the natural advantage of using LLMs over traditional deep-learning methods is their capability to generate explanations for their predictions. Here, we compare the generated explanations qualitatively between the pre-trained LLMs and our SEP model.

\begin{figure}[h]
\vspace{-1px}
\includegraphics[width=\columnwidth]{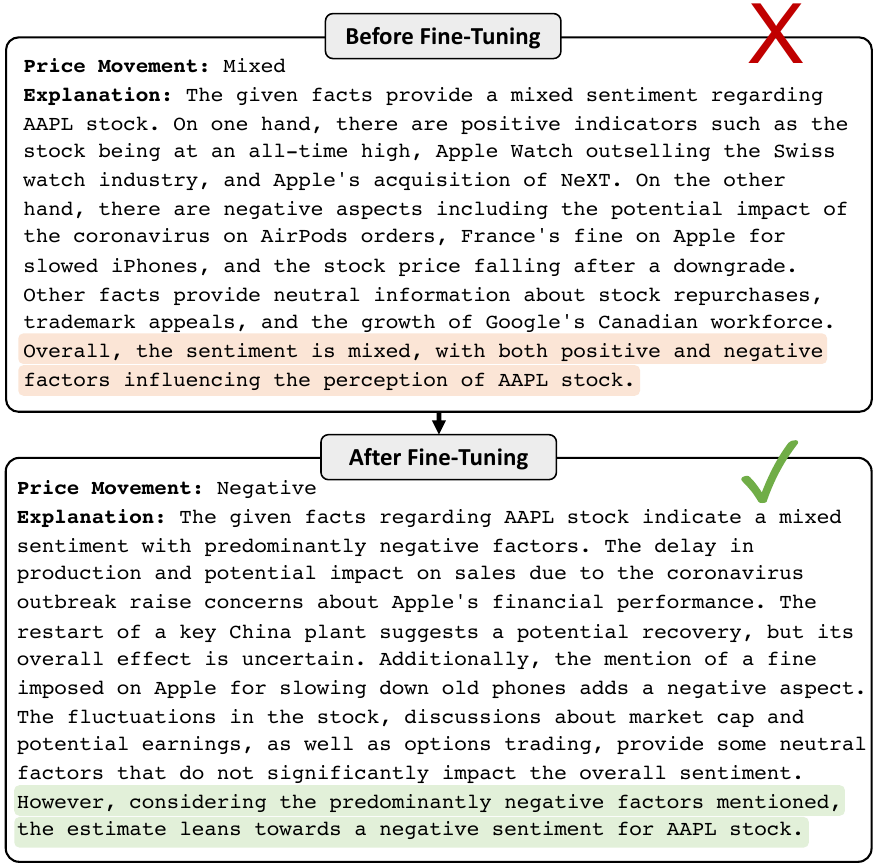}
\vspace{-22px}
\caption{An example of SEP learning to make a decisive, aggregate stock prediction. After fine-tuning, the SEP model is able to make a correct Negative prediction based on the predominantly negative events contained within the texts.}
\label{decisive}
\vspace{-22px}
\end{figure}

After SEP fine-tuning, we can observe two main improvements. The first deals with the ability to decisively weigh between news information to make a stock movement prediction. While pre-trained LLMs are known to be able to classify the sentiment of individual texts \cite{zhang2023sentiment, lopez2023can}, they typically do not try to weigh between these sentiments and make a decisive stock prediction, even if specifically requested by the prompt (see Figure \ref{uninformative}). This is generally an easier task to tackle, which is similar to fine-tuning an expert LLM \cite{guo2023close}, albeit ours is trained without human experts-in-the-loop. Figure \ref{decisive} shows an example of how our SEP model can learn how to make a decisive stock prediction after the fine-tuning process.

The second improvement deals with the ability to generate better-quality explanations. This is a more difficult task for LLMs, as it requires them to not only understand the meaning of natural language texts, but also to correctly reason out their overall impact on the stock price movement. Through the SEP framework, our LLM first learns to reason out the correct explanations via self-reflection and teach them to the PPO model, which learns to determine heuristically what is the most probable explanation at test-time. For this comparison, we came up with a set of metrics for explanation quality, and use GPT-4 to rate each response from 1 to 7 for the samples in the top-1 stocks. The average score for each metric is reported in Table \ref{quality}. Explanations of the metrics can be found in Appendix \ref{metrics}.

From Table \ref{quality}, we can make the following observations:
\vspace{-4px}
\begin{itemize}[leftmargin=*]
\item The highest scores come from more generalizable metrics, such as \textit{Consistency with Information}. Some metrics are sometimes not observable given that no such information is in the texts, which will lower their scores. However, it is fair to compare the relative scores across the LLMs, as their input texts are the same.

\item All LLMs give good-quality explanations even if the prediction is wrong, as they can naturally understand the input texts and make reasonable comparisons, but have done so incorrectly. Thus, prediction accuracy should still be the first metric to look at. 

\item Our SEP model was able to score the highest for all metrics. We note that the model was not trained on these metrics, but was only provided a reward based on correct binary predictions. Through its self-reflection and reinforcement learning, the SEP framework was able to intrinsically teach the model to better compare these factors, in order to generate better predictions. 
\end{itemize}

\begin{table}[h]\small
\caption{Comparison of explanation quality from our SEP model against baselines. The best results are boldfaced.}
\label{quality}
\vspace{-10px}
\begin{tabular}{lccc}
\toprule
\textbf{Metric}                 & \textbf{GPT-3.5} & \textbf{Vicuna} & \textbf{SEP (Ours)} \\ \midrule
Relevance to Stock Movement     & 5.407            & 5.396           & \textbf{5.449}      \\ \hline
Financial Metrics               & 2.957            & 3.146           & \textbf{3.334}      \\ \hline
Global \& Industry Factors      & 3.180            & 3.576           & \textbf{3.700}      \\ \hline
Company Developments            & 3.905            & 4.066           & \textbf{4.224}      \\ \hline
Temporal Awareness              & 3.951            & 4.066           & \textbf{4.170}      \\ \hline
Balance of Positive \& Negative & 4.030            & 4.084           & \textbf{4.224}      \\ \hline
Contextual Understanding        & 4.012            & 4.098           & \textbf{4.193}      \\ \hline
Clarity \& Coherence            & 6.271            & 6.325           & \textbf{6.439}      \\ \hline
Consistency with Information    & 5.575            & 5.652           & \textbf{6.006}      \\ \hline
Sensitivity to Updates          & 4.112            & 4.172           & \textbf{4.362}      \\
\bottomrule
\end{tabular}
\vspace{-11px}
\end{table}
\vspace{-5px}
\subsection{Ablation Study (RQ2)}
\vspace{-1px}
In this section, we evaluate the efficiency of each of the three proposed components: the Summarize, Explain and Predict modules. 
\vspace{-15px}
\subsubsection{Summarize Module}
The Summarize module reduces the noise and length of the input texts by extracting only the important, factual information. For the ablation study, we compare against the performance of using non-summarized, raw social texts in our trained SEP model. To keep the input lengths within the LLM's token limit, we try two ways of using the raw texts: Taking the 30 most shared texts and randomly sampling 30 texts, for each day.
\vspace{-8px}
\begin{table}[h]
\caption{Comparison of SEP with non-summarized and summarized input texts. The best results are boldfaced.}
\label{summarize}
\vspace{-10px}
\setlength\extrarowheight{-2px}
\begin{tabular}{lcccc}
\toprule
\textbf{}                                                                 & \multicolumn{2}{c}{\begin{tabular}[c]{@{}c@{}}Top 1 Stock\\ (GPT-3.5)\end{tabular}} & \multicolumn{2}{c}{\begin{tabular}[c]{@{}c@{}}Remaining Stocks\\ (Vicuna)\end{tabular}} \\ \cmidrule(lr){2-3} \cmidrule(lr){4-5} 
                                                                          & \textbf{Accuracy}                         & \textbf{MCC}                            & \textbf{Accuracy}                           & \textbf{MCC}                              \\ \midrule
\begin{tabular}[c]{@{}l@{}}Non-Summ.\\(Random 30)\end{tabular} & 50.75                                     & 0.0208                                  & 40.81                                       & -0.0037                                   \\ \hline
\begin{tabular}[c]{@{}l@{}}Non-Summ.\\(Top 30)\end{tabular}    & 50.81                                     & 0.0219                                  & 41.27                                       & 0.0023                                    \\ \hline
\begin{tabular}[c]{@{}l@{}}Summarized\\(All Texts)\end{tabular}           & 51.38                                     & 0.0302                                  & 47.59                                       & 0.0203                                    \\ \hline
\begin{tabular}[c]{@{}l@{}}Summarized\\(Informative)\end{tabular}   & \textbf{54.35}                            & \textbf{0.0993}                         & \textbf{50.57}                              & \textbf{0.0508}                           \\ \bottomrule
\end{tabular}
\end{table}
\vspace{-8px}
From Table \ref{summarize}, we can make the following observations:
\vspace{-2px}
\begin{itemize}[leftmargin=*]
\item Using the most shared texts is better than randomly sampling.

\item The model trained on Vicuna-generated responses fare much worse without summarizing than the GPT-3.5-trained model. This could be attributed to the texts causing more hallucination (Table \ref{hallucinationx}), given their chaotic content (\eg emojis, spam, \etc). 

\item The summarized texts provide better results. One possible reason here could be due to having information from more than 30 texts. However, it would also show that the summarization process did not lose any important information that would cause degradation.

\item Finally, removing the non-informative texts, which is only possible with the Summarize module, provides the best results.
\end{itemize}

\vspace{-5px}
\subsubsection{Explain Module}
In the SEP model, we have observed two improvements: 1) the ability to make \textit{decisive} stock predictions from mixed sentiments; and 2) the ability to make \textit{correct} stock predictions (\ie better prediction accuracy). In order to fine-tune the LLM to produce these predictions and explanations, the Explain module must first try to generate correctly-annotated samples through binary feedback and self-reflection. To demonstrate its effectiveness, we plot the percentage change in the number of generated decisive and correct predictions after each of its reflective iteration.

\begin{figure}[h]
\vspace{-8px}
\includegraphics[width=0.88\linewidth]{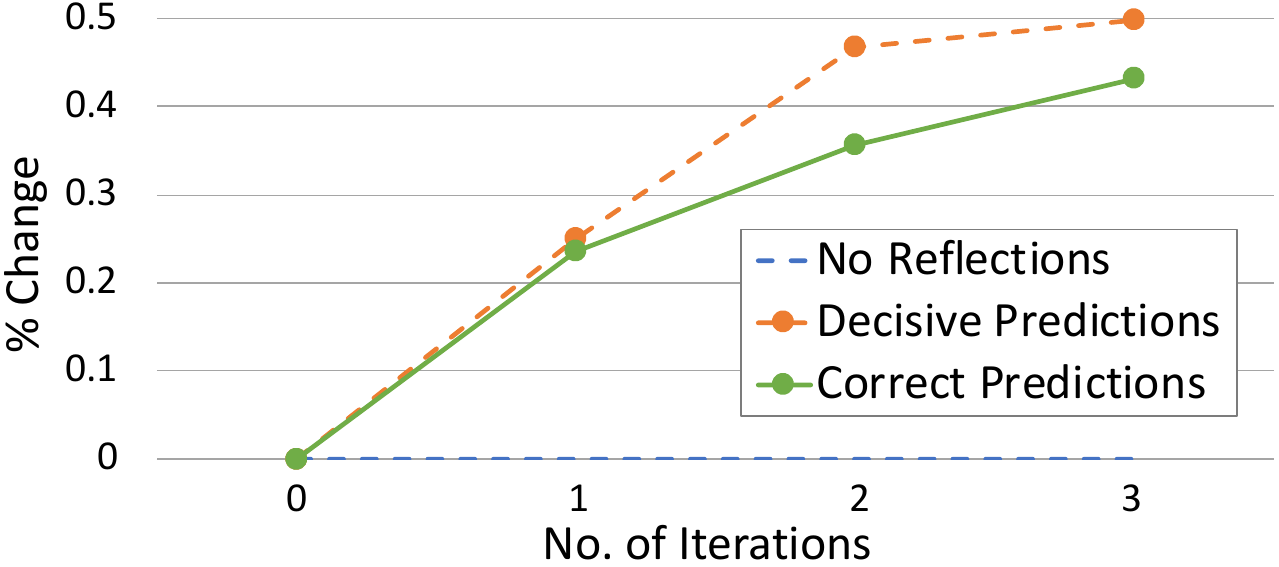}
\vspace{-11px}
\caption{Percentage change in number of decisive and correct explanation samples over the self-reflective process.}
\label{improvements}
\vspace{-10px}
\end{figure}

From Figure \ref{improvements}, we see that with multiple self-reflective iterations, the model generates more and more decisive and correct annotated samples, to be used for fine-tuning. 
We also observe that there is a greater number of decisive samples produced given that it is an easier task, which starts to slow down as more samples become non-Neutral. Overall, the number of decisive samples grew by 49.8\% while the correct samples grew by 43.2\% after 3 iterations, which highlights the effectiveness of the Explain module in generating annotated explanation samples, without the help of human experts.

\vspace{-3px}
\subsubsection{Predict Module}
For the Predict module, we conduct an ablation study over different variants of the model. We remove one additional component for each variant, \ie no $n$-shot sampling at inference \textbf{[SEP (1-shot)]}; no PPO reinforcement learning \textbf{[SEP (no PPO)]}; and no explanations \textbf{[SEP (binary)]}, which is simply instruction-tuning the LLM to make binary up/down predictions. We make the comparisons on the top-1 stock from each industry.
\begin{table}[h]
\vspace{-8px}
\caption{Ablation study. The best results are boldfaced.}
\label{ablation}
\vspace{-10px}
\begin{tabular}{lcccc}
\toprule
\textbf{}           & \multicolumn{2}{c}{All Texts}     & \multicolumn{2}{c}{Informative Texts} \\ \cmidrule(lr){2-3} \cmidrule(lr){4-5} 
                    & \textbf{Accuracy} & \textbf{MCC}    & \textbf{Accuracy}  & \textbf{MCC}    \\ \midrule
SEP (binary)        & 40.84             & -0.0042         & 42.75              & 0.0295          \\
SEP (no PPO)        & 44.08             & 0.0144          & 45.29              & 0.0368          \\
SEP (1-shot)        & 50.08             & 0.0270          & 52.54              & 0.0715          \\
SEP (Ours) & \textbf{51.38}    & \textbf{0.0302} & \textbf{54.35}     & \textbf{0.0993} \\ \bottomrule
\end{tabular}
\vspace{-8px}
\end{table}

From Table \ref{ablation}, we can make the following observations:
\vspace{-1px}
\begin{itemize}[leftmargin=*]
\item The addition of the explanation component during the instruction-tuning process, \ie from \textbf{SEP (binary)} to \textbf{SEP (no PPO)}, gives the model an average improvement of 6.9\%. It is likely that by tuning the LLM to generate explanations, we are able to elicit a reasoning process from the LLM \cite{wei2022chain} when generating stock movement predictions, resulting in better prediction accuracy. 

\item The instruction-tuned variant, \ie \textbf{SEP (no PPO)}, shows very similar results to the base model that it is tuned on (\ie the \textbf{Vicuna} model in Table \ref{results}). It is possible that the instruction tuning process has no impact on the SEP model given that the samples, taken before the reflective iterations (\ie Step 1 in Figure \ref{ppo}), are "easy" samples that the base model could already handle. We also note that supervised-tuned models have been seen to produce little to even negative improvements in previous literature \cite{stiennon2020learning}.

\item The largest improvement comes from the PPO reinforcement learning, \ie from \textbf{SEP (no PPO)} to \textbf{SEP (1-shot)}, with an average improvement of 14.8\%. This highlights the ability of the PPO trainer in teaching the LLM to generate stock predictions more effectively. Additionally, the $n$-shot sampling weighs between $n$ generated samples using the learnt reward model to select the best output. The average improvement of this variant \ie 3.0\% from \textbf{SEP (1-shot)} to \textbf{SEP (Ours)}, further reinforces the usefulness of the reward model trained during the PPO process.
\end{itemize}

\vspace{-8px}
\subsection{Portfolio Optimization (RQ3)}
\vspace{-1px}
From our results, we have observed that the SEP framework is able to teach an LLM to weigh the impact of information within the input texts in a binary manner. We further explore its generalization capability by using it to fine-tune a LLM to weigh between information within its own generated explanations quantitatively, in order to generate portfolio weights for the stock portfolio task.

For the portfolio task, we follow the same method as above to fine-tune a LLM. Here, the input information are now all the generated explanations for the basket of stocks for each day. For this experimental task, we filter only the stocks with positive predictions, in order to reduce the number of stocks the LLM have to weigh, and to prevent negative weights (hence setting a no short-sales constraint \cite{koa2023diffusion}). We then prompt the LLM to generate portfolio weights given the outlook for each given stock (see Figure \ref{portfolio}). A full example of the prompt and response can be found in Appendix \ref{prompt_samples}.

\begin{figure}[h]
\vspace{-8px}
\includegraphics[width=\columnwidth]{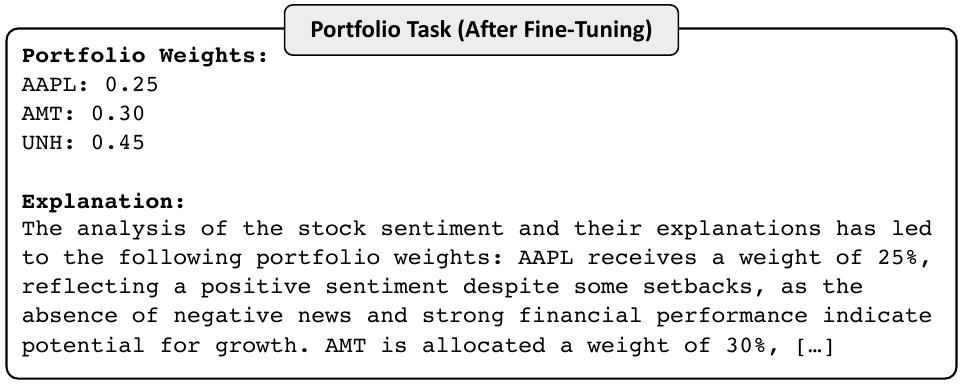}
\vspace{-21px}
\caption{An example response for the portfolio task after fine-tuning with SEP. [...] refers to truncated text.}
\label{portfolio}
\vspace{-9px}
\end{figure}

As there is no binary feedback for this task, in each self-reflective iteration, we provide the reflective LLM with the overall profits based on the provided portfolio weights, and prompt it to reflect on how it can improve itself to obtain higher profits. The reflections are then used to generate an updated set of portfolio weights. Finally, we feed both sets of generated weights into a PPO trainer, where the one with higher profits is used as the "better" response. 

We compare the performances of portfolios generated by three different LLMs: GPT-3.5-turbo, Vicuna, and our fine-tuned SEP model. We also include three baselines: the 1/N portfolio, where all 11 stocks in the basket are bought at equal weights \cite{demiguel2009optimal}; the S\&P500 stock market index; and Positive-Only, where only the predicted positive stocks are bought at equal weights. The latter can also be seen as evaluating the results of the original stock prediction LLM in a practical setting, without the portfolio weighing prompts.  

We evaluate the portfolio performance using four metrics: the overall gain, which simply sums up the gains for each day; the cumulative gain, which is the final gain after re-investing any additional profits or losses over the evaluation period; the standard deviation of the profits; and the annualized Sharpe Ratio \cite{lo2002statistics}. 
\begin{table}[h]
\vspace{-11px}
\caption{Portfolio results comparison. The best results are boldfaced. The Sharpe Ratio values are annualized.}
\vspace{-12px}
\label{sharpe}
\begin{tabular}{|l|c|c|c|c|}
\hline
\multicolumn{1}{|l|}{\textbf{Approach}} & \textbf{Overall} & \textbf{Cumulative} & \textbf{Std. Dev.} & \textbf{Sharpe} \\ \hline
1/N                                     & -0.0330                                                          & -0.0502                                                             & 1.613e-2                                                              & -0.225                                                          \\ \hline
Market Index                                   & 0.0180                                                           & 0.0003                                                              & \textbf{1.533e-2}                                                              & 0.123                                                           \\ \hline
Positive-Only                                & 0.1243                                                           & 0.1065                                                              & 1.911e-2                                                              & 0.807                                                           \\ \hline
GPT-3.5                                 & 0.1497                                                           & 0.1353                                                              & 1.893e-2                                                              & 0.980                                                           \\ \hline
Vicuna                                  & 0.1541                                                           & 0.1447                                                              & 1.731e-2                                                              & 1.104                                                           \\ \hline
SEP (Ours)                     & \textbf{0.1661}                                                  & \textbf{0.1569}                                                     & 1.792e-2                                                     & \textbf{1.150}                                                  \\ \hline
\end{tabular}
\vspace{-15px}
\end{table}

Table \ref{sharpe} reports the portfolio results. From the table, we observe:

\begin{itemize}[leftmargin=*]
\item The \textbf{Positive-Only} portfolio, \ie evenly buying the stocks that are predicted to be Positive, already showcases good performance. This highlights the capability of our original stock prediction model to produce good trading signals in a practical setting.

\item For the standard deviation results, we note that the top 2 portfolio methods, \ie \textbf{1/N} and \textbf{Market Index}, contains more number of stocks, which allow them to spread out the stock price fluctuations more evenly. However, their Sharpe Ratios are still lower than the other models, which shows a lower reward-to-risk ratio.

\item The pre-trained LLM models, \ie \textbf{GPT-3.5} and \textbf{Vicuna}, already shows better performance than the \textbf{Positive-Only} portfolio in most metrics, which shows the capabilities of using LLMs to weigh between information to produce portfolio weights.

\item Our \textbf{SEP} model was able to outperform all other methods in most portfolio metrics, 
and achieve comparable performance in its standard deviation, 
which showcases the effectiveness of our SEP framework. 
In addition to the shown metrics, we also re-emphasize the ability of the LLM-based models to \textit{explain} the generated portfolio weights, which further adds to the interpretability and trustability of their results for practitioners.
\end{itemize}

\vspace{-8px}
\section{Conclusion and Future Work}
\vspace{-2px}
In this work, we explored the explainable stock prediction task, which was largely difficult to solve before generative models. For this task, we highlighted two challenges: the limitations of current LLMs in weighing varied market factors to make aggregate stock predictions, and the lack of annotated training samples for fine-tuning LLMs to make explanations. To tackle these challenges, we proposed our SEP framework, which utilizes a verbal self-reflective agent and PPO techniques to let a LLM teach itself how to generate stock explanations in a fully autonomous manner. Through experimental results, we validated that our SEP model is able to outperform both traditional deep-learning and LLM methods in the accuracy of the predictions and quality of the generated explanations. Furthermore, we also demonstrated the generalizability of the SEP framework by fine-tuning a model for the portfolio task. 


There are some directions that can be explored in future works. Firstly, we address the possibility of cumulative errors in the SEP framework. At each stage, poorly generated summaries or explanations could lead to poorer responses in the next step. In practice, it is possible for experts to vet through the responses before using them, which would be an easier task than generating them manually. However, more can be done to increase the robustness of the generated responses and reduce the need for human-in-the-loop. Secondly, using additional data sources, such as knowledge graphs \cite{hansen2023can} or audio features \cite{yang2022numhtml}, could increase the quality of the predictions. At the same time, such works would also help to explore the multi-modal capabilities of the most recent LLM upgrades \cite{bang2023multitask, wu2023next}. 
Finally, as this is a relatively new task, there are currently limited works on evaluating the generated stock explanations. Further studies can be done to improve the metrics created in this work. 
\section{Ethical Use of Data}
\vspace{-2px}
For this research, we have utilized datasets derived from publicly available sources, and no human annotators were involved in the data collection process. Rights pertaining to the data used, such as text data, remain the sole property of the original rights holders. 
This study is intended exclusively for academic purposes only. 

There are potential ethical and social implications of using LLMs for stock prediction. We list some of them here and suggest possible ways to mitigate them when deploying our model for practical use:
\begin{itemize}[leftmargin=*]
\vspace{-2px}
\item \textit{Risk of Manipulation.}
Market manipulation has always been a problem in stock markets \cite{li2023detecting}. Using LLMs for stock prediction can increase this risk, given their known vulnerabilities such as jailbreak prompting \cite{wei2023jailbroken} and model red-teaming \cite{casper2023explore}. To mitigate these, there should be measures to scrutinize user inputs to the model before processing them. Access to the LLM’s internal knowledge base should be restricted to authorized users only.

\item \textit{Misinformation.}
While the point of explainable forecasting is to generate trustable results, LLMs can also be leveraged to generate deceptive misinformation \cite{chen2023combating}. Measures should be taken to verify the correctness of facts before utilizing them in the explanations, either by automated verification \cite{zhang2023towards} or human-in-the-loop.

\item \textit{Prediction Bias.}
It is known that LLMs tend to inherit stereotypes or existing biases due to the internet-based data they are trained on \cite{gallegos2023bias}. As stock prediction with LLMs is relatively new, it is unknown whether existing investor biases \cite{niessen2019sex, jannati2023group} will also carry over into the LLMs’ generated responses. 
Some mitigation strategies include removing biased responses, verifying all information are factual, before training the LLM with reinforcement learning. 
\vspace{-12px}
\end{itemize}

In general, the most effective mitigation strategy is to include human-in-the-loop to anticipate and mitigate various potential risks. While LLMs can assist humans in labor-intensive tasks such as processing large volume of texts and analyzing their stock market impacts, it is not able to replace the need for human oversight.
\vspace{-5px}
\vspace{-1px}
\section{Acknowledgement}
\vspace{-2px}
This research is supported by the National Research Foundation, Singapore under its Industry Alignment Fund – Pre-Positioning (IAF-PP) Funding Initiative, by the National Research Foundation, Singapore through the National Cybersecurity R\&D Lab at the National University of Singapore under its National Cybersecurity R\&D Programme (Award No. NCR25-NCL P3-0001). Any opinions, findings and conclusions or recommendations expressed in this material are those of the author(s) and do not reflect the views of National Research Foundation, Singapore.
\vspace{-5px}

\bibliographystyle{ACM-Reference-Format}
\bibliography{reference}

\appendix
\newpage
\vspace{-6px}
\section{Dataset and Clustering Pipeline}\label{datasets}
\vspace{-2px}
In this section, we include additional details on the statistics of the collected dataset and the overall clustering pipeline.

\vspace{-9px}
\subsection{Dataset}
\vspace{-2px}
In this work, we construct a new dataset by following the data collection methodology used for the \textbf{ACL18} StockNet dataset \cite{xu2018stock}, updated for the year 2020–2022 (see Table \ref{dataset-stocks} for a list of included stock companies). Since the previous work, the number of tweets have increased exponentially
(see Table \ref{cluster-stats}). To keep the most relevant texts within a reasonable length, we first employ a clustering pipeline to obtain the most representative tweets for each day.

\vspace{-9px}
\subsection{Clustering Pipeline}
\vspace{-2px}
Following previous works that perform clustering on full-length documents for LLM inputs \cite{ma2023structured}, we make use of the BERTopic \cite{grootendorst2022bertopic} pipeline for clustering: First, we generate embeddings for the tweets using a pre-trained language model RoBERTa \cite{liu2019roberta}, which have also been fine-tuned using the SimCSE \cite{gao2021simcse} framework. Next, UMAP \cite{mcinnes2018umap} was used for dimensionality reduction of the embeddings, and HDBSCAN \cite{campello2013density} was used to cluster them into semantically similar groups. Finally, we use a class-based TF-IDF procedure \cite{grootendorst2022bertopic, rajaraman2011mining} to rank and extract the most representative tweet for each cluster. 

For the hyper-parameters, we set the number of neighbors for UMAP dimensionality reduction as 15. For HDBSCAN clustering, the minimum cluster size is set to 10. 
The statistics of the tweet data before and after clustering can be found in Table \ref{cluster-stats}. 

\begin{table}[h]
\vspace{-12px}
\caption{Statistics of tweets before and after clustering.}
\vspace{-10px}
\label{cluster-stats}
\begin{tabular}{lcccc}
\hline
                  & \textbf{\begin{tabular}[c]{@{}c@{}}Avg.\\ tweets\end{tabular}} & \textbf{\begin{tabular}[c]{@{}c@{}}Avg.\\ tokens\end{tabular}} & \textbf{\begin{tabular}[c]{@{}c@{}}Max\\ tweets\end{tabular}} & \textbf{\begin{tabular}[c]{@{}c@{}}Max\\ tokens\end{tabular}} \\ \hline
Before Clustering & 469                                                            & 27,951                                                         & 46,569                                                        & 1,911,495                                                     \\ \hline
After Clustering  & 16                                                             & 1,068                                                          & 1,599                                                         & 63,392                                                        \\ \hline
\end{tabular}
\vspace{-12px}
\end{table}

In total, the dataset consists of tweets for 757 trading days. The overall number of samples used is 29,997, which is split in a train-test ratio of 8:2. Within the training set, 10\% of the generated explanation samples are used for validation during fine-tuning.

\begin{table*}[!htbp]\small

\caption{Top 5 stocks and their companies selected from the 11 industries.}

\label{dataset-stocks}
\begin{tabular}{l|l|l}
\hline
\textbf{Sector}                          & \textbf{Stock symbol} & \textbf{Company}                                   \\ \hline
\multirow{5}{*}{Basic Materials}        & \$BHP                 & BHP Group Limited                                  \\
                                         & \$RIO                 & Rio Tinto Group                                    \\
                                         & \$SHW                 & The Sherwin-Williams Company                       \\
                                         & \$VALE                & Vale S.A.                                          \\
                                         & \$APD                 & Air Products and Chemicals, Inc.                   \\
 \hline\hline
\multirow{5}{*}{Financial Services}     & \$BRK-A               & Berkshire Hathaway Inc.                            \\
                                         & \$V                   & Visa Inc.                                          \\
                                         & \$JPM                 & JPMorgan Chase \& Co.                              \\
                                         & \$MA                  & Mastercard Inc.                                    \\
                                         & \$BAC                 & Bank of America Corporation                        \\
 \hline\hline
\multirow{5}{*}{Consumer Defensive\quad\quad}     & \$WMT                 & Walmart Inc.                                       \\
                                         & \$PG                  & The Procter \& Gamble Company                      \\
                                         & \$KO                  & The Coca-Cola Company                              \\
                                         & \$PEP                 & PepsiCo, Inc.                                      \\
                                         & \$COST                & Costco Wholesale Corporation                       \\
 \hline\hline
\multirow{5}{*}{Utilities}              & \$NEE                 & NextEra Energy, Inc.                               \\
                                         & \$DUK                 & Duke Energy Corporation                            \\
                                         & \$SO                  & The Southern Company                               \\
                                         & \$D                   & Dominion Energy, Inc.                              \\
                                         & \$AEP                 & American Electric Power Company, Inc.\quad\quad\quad\quad\quad\quad\quad\quad\quad\quad              \\
 \hline\hline
\multirow{5}{*}{Energy}                 & \$XOM                 & Exxon Mobil Corporation                            \\
                                         & \$CVX                 & Chevron Corporation                                \\
                                         & \$SHEL                & Shell plc                                          \\
                                         & \$TTE                 & TotalEnergies SE                                   \\
                                         & \$COP                 & ConocoPhillips                                     \\
\hline\hline                                        
\multirow{5}{*}{Technology}             & \$AAPL                & Apple Inc.                                         \\
                                         & \$MSFT                & Microsoft Corporation                              \\
                                         & \$TSM                 & Taiwan Semiconductor Manufacturing Company Limited \\
                                         & \$NVDA                & NVIDIA Corporation                                 \\
                                         & \$AVGO                & Broadcom Inc.                                      \\
\hline\hline
\multirow{5}{*}{Consumer Cyclical}      & \$AMZN                & Amazon.com, Inc.                                   \\
                                         & \$TSLA                & Tesla, Inc.                                        \\
                                         & \$HD                  & The Home Depot, Inc.                               \\
                                         & \$BABA                & Alibaba Group Holding Limited                      \\
                                         & \$TM                  & Toyota Motor Corporation                           \\
 \hline\hline
 \multirow{5}{*}{Real Estate}            & \$AMT                 & American Tower Corporation                         \\
                                         & \$PLD                 & Prologis, Inc.                                     \\
                                         & \$CCI                 & Crown Castle Inc.                                  \\
                                         & \$EQIX                & Equinix, Inc.                                      \\
                                         & \$PSA                 & Public Storage                                     \\
 \hline\hline
\multirow{5}{*}{Healthcare}             & \$UNH                 & UnitedHealth Group Incorporated                    \\
                                         & \$JNJ                 & Johnson \& Johnson                                 \\
                                         & \$LLY                 & Eli Lilly and Company                              \\
                                         & \$PFE                 & Pfizer Inc.                                        \\
                                         & \$ABBV                & AbbVie Inc.                                        \\
  \hline\hline
\multirow{5}{*}{Communication Services} & \$GOOG                & Alphabet Inc.                                      \\
                                         & \$META                & Meta Platforms, Inc.                               \\
                                          & \$VZ                  & Verizon Communications Inc.                        \\
                                        & \$CMCSA\quad\quad\quad                & Comcast Corporation                                \\
                                         & \$DIS                 & The Walt Disney Company                            \\
 \hline\hline
\multirow{5}{*}{Industrials\quad\quad\quad\quad\quad\quad\quad }             & \$UPS                  & United Parcel Service, Inc.                                   \\
                                         & \$UNP\quad\quad\quad\quad\quad                 & Union Pacific Corporation                    \\
                                         & \$HON                  & Honeywell International Inc.                                 \\
                                         & \$LMT                 & Lockheed Martin Corporation Company\quad\quad\quad\quad\quad\quad\quad\quad\quad\quad                  \\
                                         & \$CAT                 & Caterpillar Inc.                  \\ \hline


\end{tabular}
\end{table*}

\vspace{-5px}
\section{Full Prompt Examples}\label{prompt_samples}
\vspace{-2px}
In this section, we provide full examples of the prompts used in SEP and the responses. Examples for four tasks are shown:

\begin{itemize}[leftmargin=*]
\vspace{1px}
\item Table \ref{summary_full} shows an example for the summarization task, where summarized factual information is generated from the chaotic input tweets. In the example, we can see that tweets that contain useless information, such as unsubstantiated comments, are ignored by the LLM. Additionally, the facts extracted from the tweets are also summarized in a concise and succinct manner. 

\vspace{1px}
\item Table \ref{explanation_full} shows a successful example for the explanation task. In the example, we can see that while there are some positive news, there are more recent and impactful negative facts which caused a negative price movement. The example showcases the ability of the LLM to weigh between these factors effectively, and generate the correct price movement with a reasonable explanation.

\vspace{1px}
\item Table \ref{reflection_full} shows an example for the reflection task. In the example, the incorrect previous response is fed into the LLM to generate a reflection, which consists of what went wrong and a plan on how to mitigate this problem. The reflection tells the LLM to further consider the positive earnings, overall market for big tech companies, and the long-term strategic initiatives, which allowed it to obtain a correct prediction in the next iteration.

\vspace{1px}
\item Table \ref{portfolio_full} shows an example for the portfolio task. Given the self-predicted explanations for all positive stocks for each day, the LLM further weigh between their outlook to recommend the amount of each stock to purchase. In the example, we can see the LLM gave more weight to factors such as digital transformation, which could signify potential future growth for the company. 
\end{itemize}

\begin{table*}[htbp]
\caption{A full prompt and its response for generating summarized facts. Underlined words denote the end of the input text. 
}
\label{summary_full}
\vspace{-42px}
  \centering
  \makebox[\textwidth]{
  \begin{tabular}{ccc}   
   \includegraphics{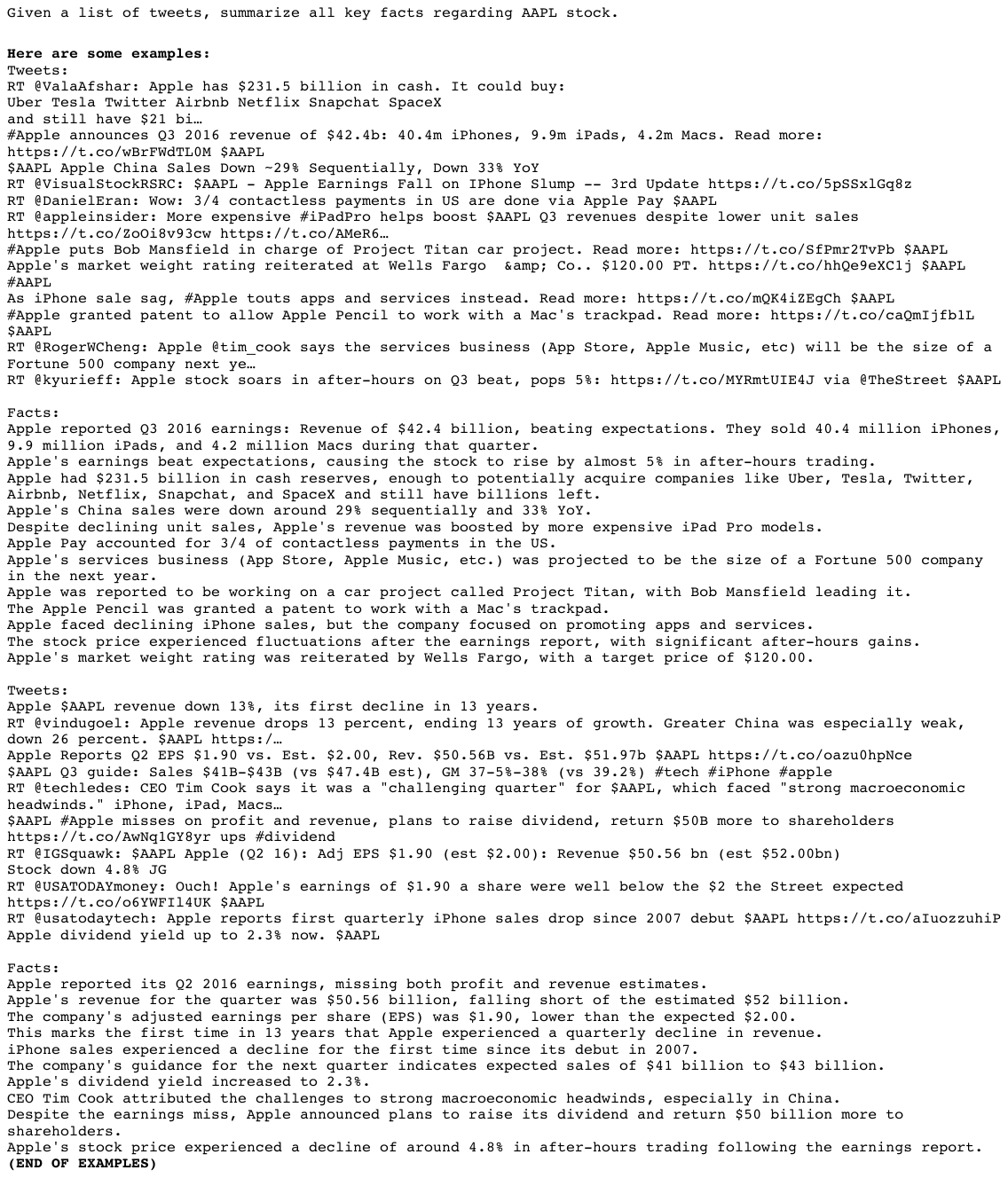}          
  \end{tabular}}
\end{table*}
\begin{table*}[htbp]\ContinuedFloat
\vspace{-27px}
  \centering
  \makebox[\textwidth]{
  \begin{tabular}{ccc}   
   \includegraphics{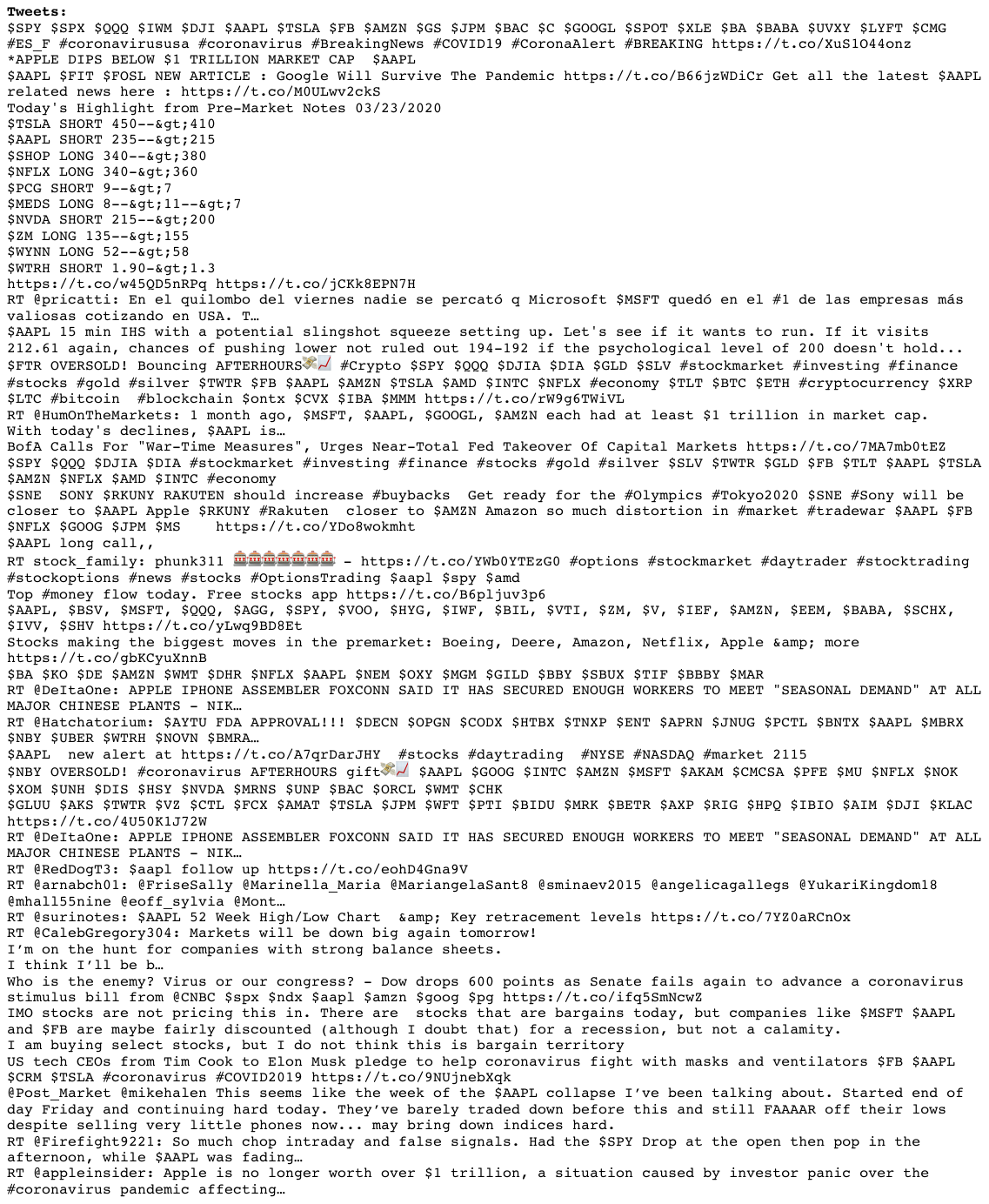}          
  \end{tabular}}
\end{table*}
\begin{table*}[htbp]\ContinuedFloat
\vspace{-27px}
  \centering
  \makebox[\textwidth]{
  \begin{tabular}{ccc}   
   \includegraphics{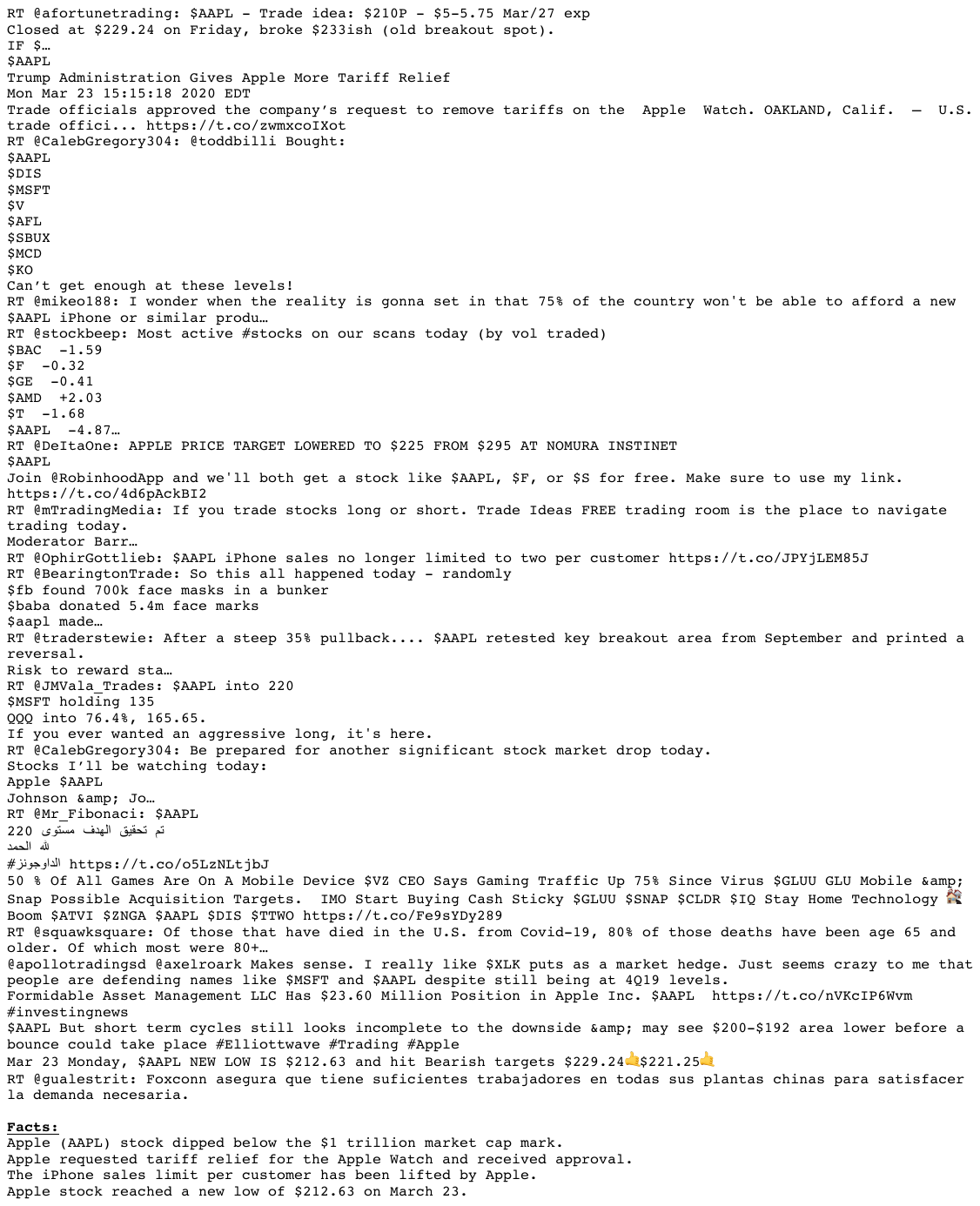}          
  \end{tabular}}
\end{table*}

\begin{table*}[htbp]
\caption{A full prompt and its response for generating stock explanations. Underlined words denote the end of the input text.}
\label{explanation_full}
\vspace{-30px}
  \centering
  \makebox[\textwidth]{
  \begin{tabular}{ccc}   
   \includegraphics{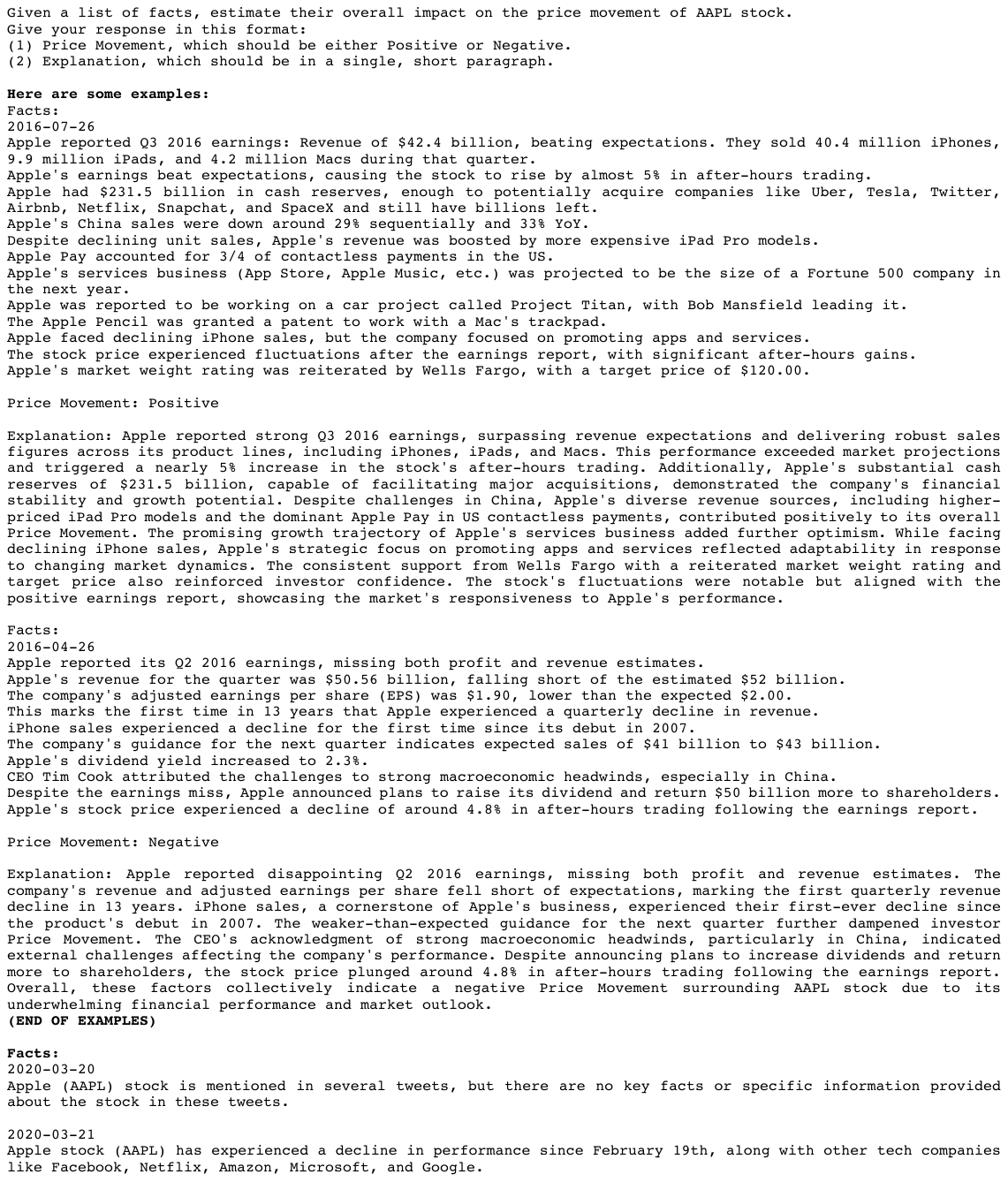}          
  \end{tabular}}
\end{table*}
\begin{table*}[htbp]\ContinuedFloat
\vspace{-27px}
  \centering
  \makebox[\textwidth]{
  \begin{tabular}{ccc}   
   \includegraphics{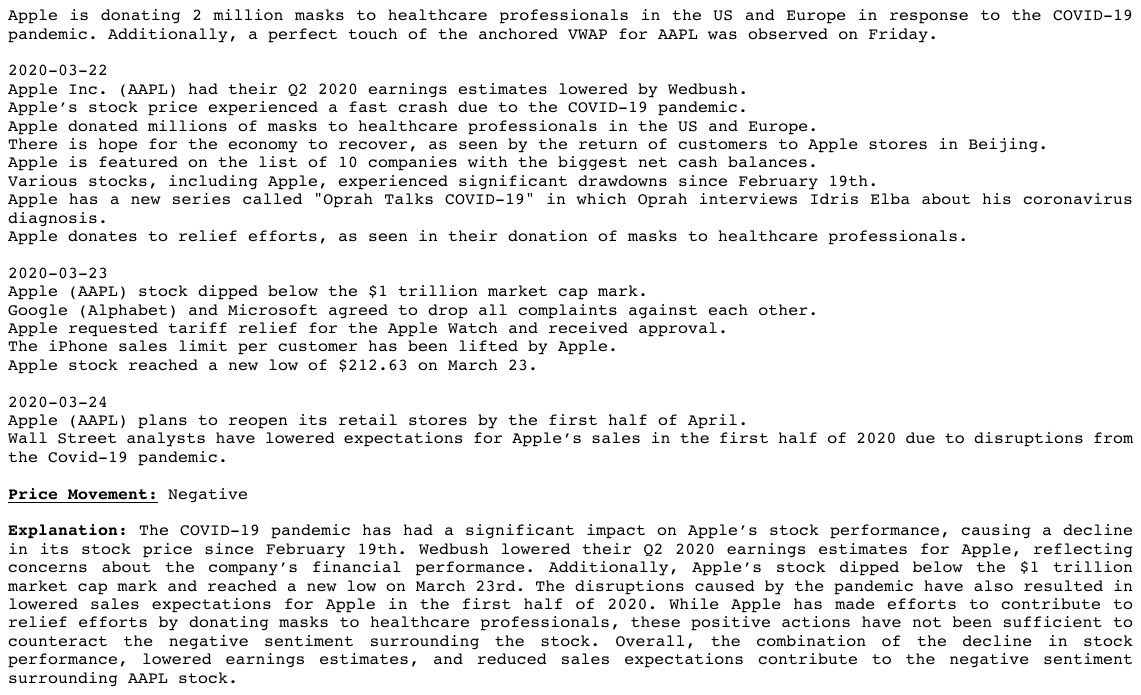}          
  \end{tabular}}
\end{table*}

\begin{table*}[htbp]
\caption{A full prompt and its response for generating reflections. Underlined words denote the end of the input text.}
\label{reflection_full}
\vspace{-30px}
  \centering
  \makebox[\textwidth]{
  \begin{tabular}{ccc}   
   \includegraphics{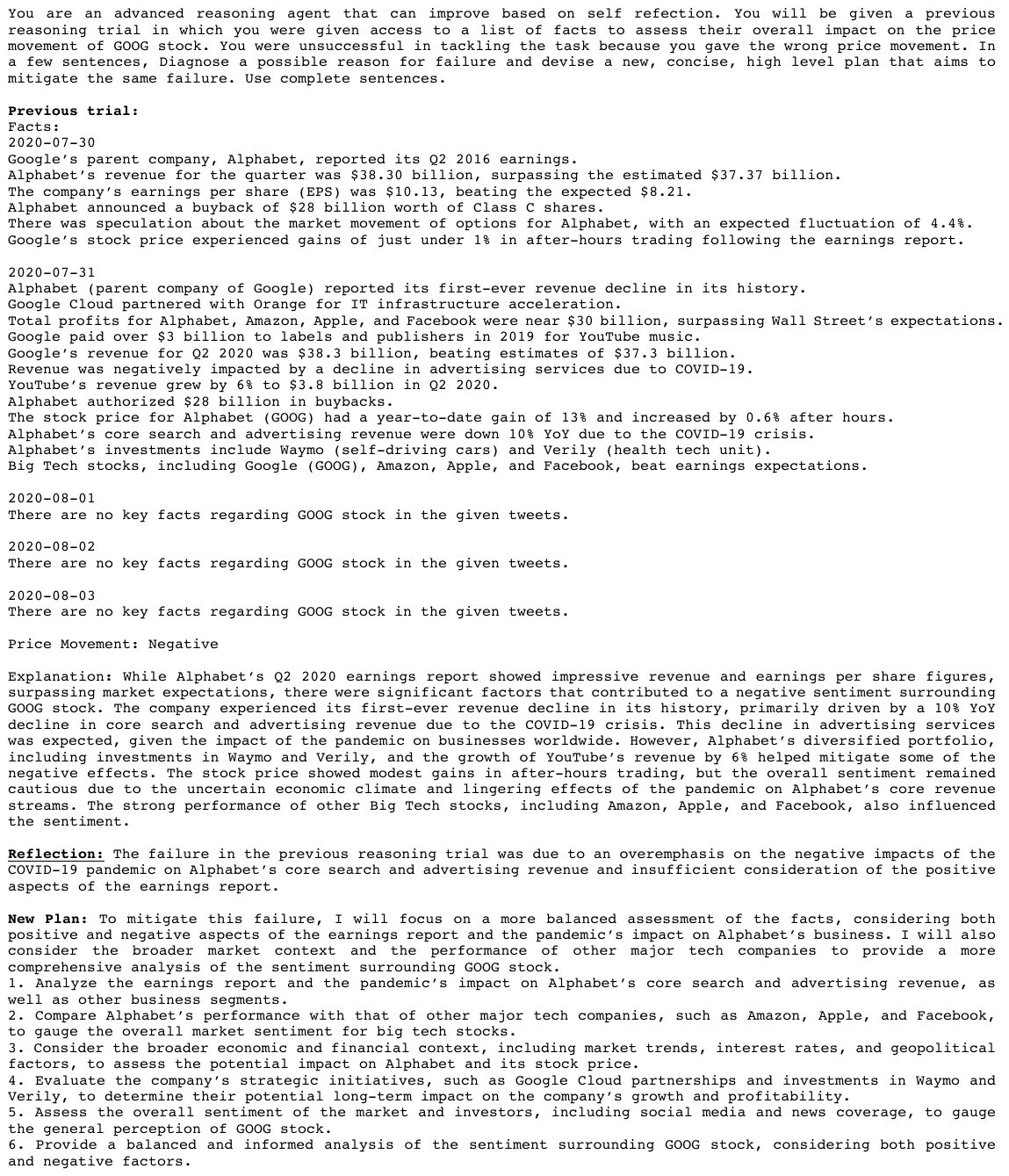}          
  \end{tabular}}
\end{table*}

\begin{table*}[htbp]
\caption{A full prompt and its response for generating portfolio weights. Underlined words denote the end of the input text.}
\label{portfolio_full}
\vspace{-30px}
  \centering
  \makebox[\textwidth]{
  \begin{tabular}{ccc}   
   \includegraphics{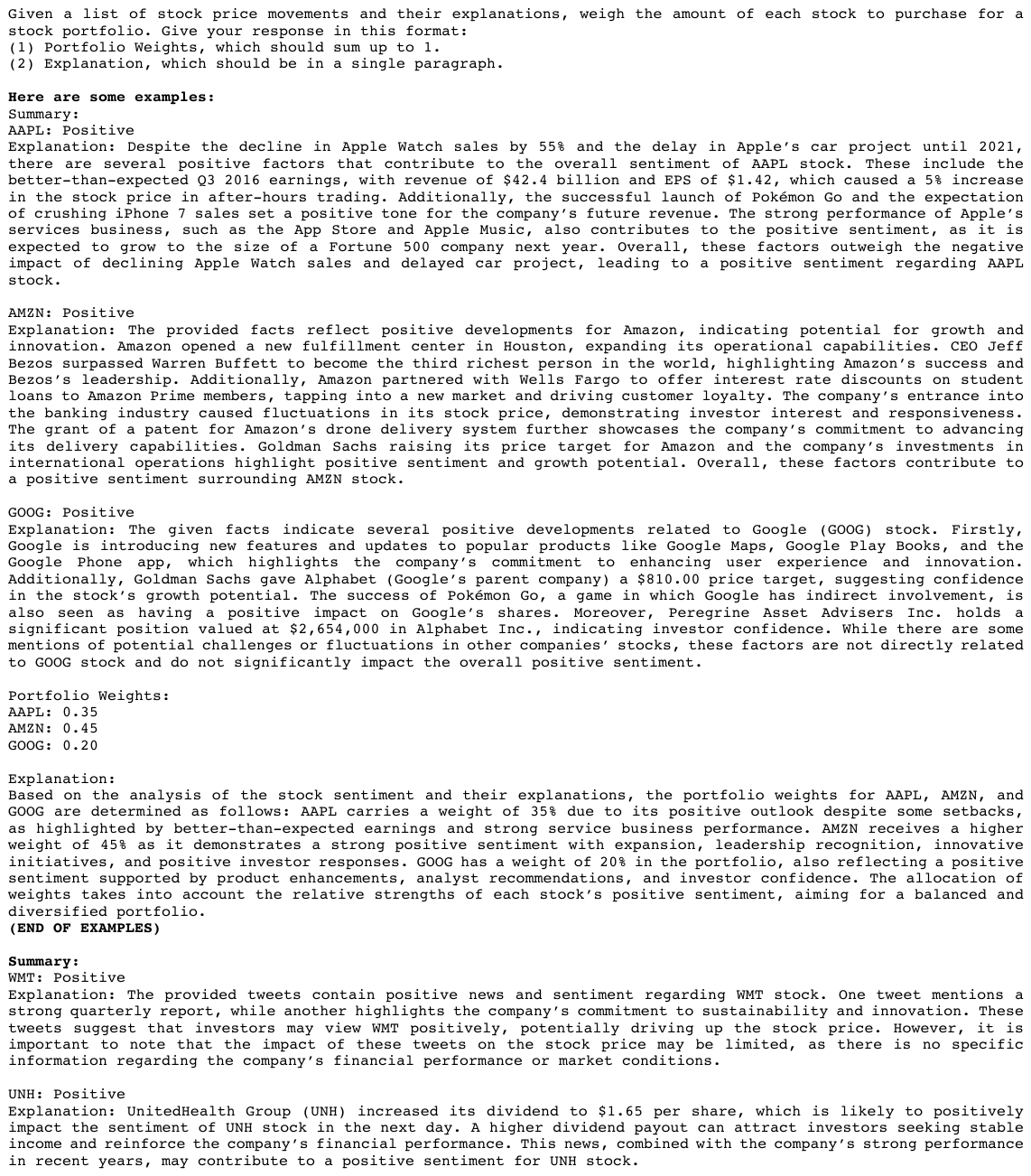}          
  \end{tabular}}
\end{table*}
\begin{table*}[htbp]\ContinuedFloat
\vspace{-27px}
  \centering
  \makebox[\textwidth]{
  \begin{tabular}{ccc}   
   \includegraphics{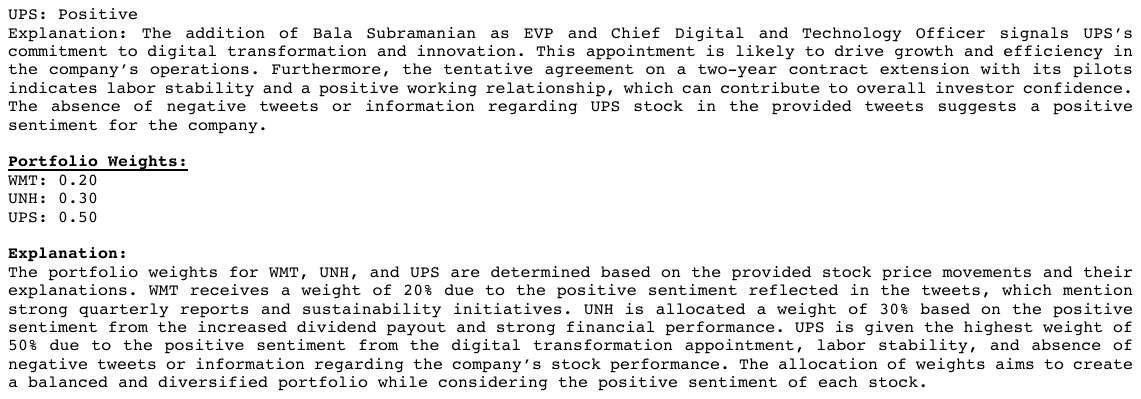}          
  \end{tabular}}
\end{table*}

\vspace{-5px}
\section{Evaluation of Explanation Quality}\label{metrics}
\vspace{-2px}
The explanation of the metrics used in Table \ref{quality} are given below. These metrics are manually curated by us through the assistance of ChatGPT. There are currently limited works on evaluating generated stock explanations, given that it is a relatively new application. 

\begin{itemize}[leftmargin=*]
\item \textbf{Relevance to Stock Movement:}
\begin{itemize}
\item Does the explanation focus on factors directly related to the stock's movement?
\end{itemize}

\item \textbf{Financial Metrics:}
\begin{itemize}
\item Does the explanation include relevant financial metrics (e.g., earnings estimates, market cap)?
\item Are these metrics explained in the context of their impact on stock performance?
\end{itemize}

\item \textbf{Global \& Industry Factors:}
\begin{itemize}
\item Does the explanation consider broader economic conditions or industry trends that may impact the stock?
\item Is there an understanding of how global events could influence the stock's performance?
\end{itemize}

\item \textbf{Company Developments:}
\begin{itemize}
\item Are specific developments related to the company discussed?
\item Is there an understanding of how these developments might influence the stock?
\end{itemize}

\item \textbf{Temporal Awareness:}
\begin{itemize}
\item Does the explanation consider the timing of events and developments?
\item Is there an acknowledgment of the temporal dynamics of the stock market?
\end{itemize}

\item \textbf{Balance of Positive \& Negative:}
\begin{itemize}
\item Is there an attempt to balance positive and negative factors?
\item Does the explanation recognize mitigating factors that could counteract positive or negative sentiments?
\end{itemize}

\item \textbf{Contextual Understanding:}
\begin{itemize}
\item Does the explanation demonstrate a nuanced understanding of the context in which the news is presented?
\item Is there an awareness of the complexities and uncertainties in predicting stock movements?
\end{itemize}

\item \textbf{Clarity \& Coherence:}
\begin{itemize}
\item Is the explanation clear and easy to understand?
\item Does it present a coherent narrative that connects various factors logically?
\end{itemize}

\item \textbf{Consistency with Information:}
\begin{itemize}
\item Is the information consistent with known facts and events?
\item Are there any inaccuracies or contradictions in the explanation?
\end{itemize}

\item \textbf{Sensitivity to Updates:}
\begin{itemize}
\item Does the explanation show sensitivity to the possibility of changing circumstances or new information that could affect the stock?
\end{itemize}
\end{itemize}

\end{document}